%% file: acl_latex.tex
\pdfoutput=1

\documentclass[11pt]{article}

\usepackage[final]{acl}

\usepackage{times}
\usepackage{latexsym}

\usepackage[T1]{fontenc}

\usepackage[utf8]{inputenc}

\usepackage{microtype}

\usepackage{inconsolata}
\usepackage{listings}

\usepackage{graphicx}

\usepackage{xcolor}  
\usepackage{pxrubrica}        
\usepackage{url}
\usepackage{booktabs}
\usepackage{multirow}
\usepackage{xspace}
\usepackage{amsmath}
\usepackage{amssymb}
\usepackage{subfigure}
\usepackage{comment}
\usepackage{makecell}
\usepackage{color}
\usepackage{pifont}
\usepackage{tabularx}
\usepackage{colortbl}
\usepackage{listings}
\usepackage{caption}
\usepackage{enumitem}
\usepackage{courier}
\usepackage{adjustbox}

\definecolor{red}{rgb}{0.8,0,0}
\definecolor{green}{rgb}{0,0.8,0}
\definecolor{blue}{rgb}{0,0,0.8}
\definecolor{yellow}{rgb}{0.90,0.91,0.75}
\definecolor{purple}{rgb}{0.85,0.80,0.91}

\definecolor{codegreen}{rgb}{0,0.6,0}
\definecolor{codegray}{rgb}{0.5,0.5,0.5}
\definecolor{codepurple}{rgb}{0.58,0,0.82}
\definecolor{codeblue}{rgb}{0.25,0.5,0.75}
\definecolor{backcolour}{rgb}{0.95,0.95,0.92}
\definecolor{codeorange}{rgb}{1,0.6,0}
\definecolor{codeyellow}{rgb}{0.7,0.7,0}
\definecolor{codeteal}{rgb}{0,0.5,0.5}

\lstdefinestyle{mystyle}{
    backgroundcolor=\color{backcolour},
    commentstyle=\color{codegreen},
    keywordstyle=\color{codeblue},
    numberstyle=\tiny\color{black}, 
    stringstyle=\color{black}, 
    basicstyle=\footnotesize\ttfamily,
    breakatwhitespace=false,
    breaklines=true,
    captionpos=b,
    keepspaces=true,
    numbers=left,
    numbersep=5pt,
    showspaces=false,
    showstringspaces=false,
    showtabs=false,
    tabsize=2,
    morekeywords={null,true,false},
    keywordstyle=\color{codeblue}, 
    morestring=[b]",
    stringstyle=\color{black}, 
    literate=
     *{0}{{{\color{black}0}}}{1}
      {1}{{{\color{black}1}}}{1}
      {2}{{{\color{black}2}}}{1}
      {3}{{{\color{black}3}}}{1}
      {4}{{{\color{black}4}}}{1}
      {5}{{{\color{black}5}}}{1}
      {6}{{{\color{black}6}}}{1}
      {7}{{{\color{black}7}}}{1}
      {8}{{{\color{black}8}}}{1}
      {9}{{{\color{black}9}}}{1}
      {"}{{{\color{black}"}}}{1} 
      {:}{{{\color{codeblue}:}}}{1} 
      {,}{{{\color{black},}}}{1} 
      {\{}{{{\color{codeblue}\{}}}{1} 
      {\}}{{{\color{codeblue}\}}}}{1} 
      {[}{{{\color{codeblue}[}}}{1} 
      {]}{{{\color{codeblue}]}}}{1}, 
}

\newcommand{\datasetName}[0]{LegalViz\xspace}
\renewcommand{\paragraph}[1]{\noindent\textbf{#1}.\hspace{2pt}}

\title{LegalViz: Legal Text Visualization by Text To Diagram Generation}

\author{
    Eri Onami$^{1,2}$   
    Taiki Miyanishi$^{3,6,2}$ 
    Koki Maeda$^{4,7}$ 
    Shuhei Kurita$^{5,7,2}$\\
      $^{1}$Nara Institute of Scient and Technology 
      $^{2}$RIKEN AIP 
      $^{3}$The University of Tokyo \\
      $^{4}$Institute of Science Tokyo 
      $^{5}$National Institution of Informatics
      $^{6}$ATR
      $^{7}$NII LLMC\\
      \texttt{onami.eri.ob6@is.naist.jp},
      \texttt{taiki.miyanishi@weblab.t.u-tokyo.ac.jp},\\
      \texttt{koki.maeda@nlp.c.titech.ac.jp},
      \texttt{skurita@nii.ac.jp} \\}

\begin{document}
\maketitle
\begin{abstract}
Legal documents including judgments and court orders require highly sophisticated legal knowledge for understanding.
To disclose expert knowledge for non-experts, we explore the problem of visualizing legal texts with easy-to-understand diagrams and propose a novel dataset of \datasetName with 23 languages and 7,010 cases of legal document and visualization pairs, using the DOT graph description language of Graphviz.
\datasetName provides a simple diagram from a complicated legal corpus identifying legal entities, transactions, legal sources, and statements at a glance, that are essential in each judgment.
In addition, we provide new evaluation metrics for the legal diagram visualization by considering graph structures, textual similarities, and legal contents.
We conducted empirical studies on few-shot and finetuning large language models for generating legal diagrams and evaluated them with these metrics, including legal content-based evaluation within 23 languages. Models trained with LegalViz outperform existing models including GPTs, confirming the effectiveness of our dataset.
\end{abstract}

\section{Introduction}
Driven by the rapid advancements in large language model (LLM) performance~\citep{NEURIPS2020_1457c0d6, OpenAI2023GPT4TR}, 
adaptation to specialized domains in Natural Language Processing (NLP) receives increasing attention in many fields~\citep{lu2022learn,kung2023performance,legalbench2023}.
Specifically, the application of LLMs to the legal field holds the potential to automate significant tasks and support roles traditionally occupied by lawyers and judges~\citep{choi2021chatgpt,frankenreiter2022natural}.
The understanding of legal documents poses unique challenges for NLP applications, as legal reasoning requires not only interpreting the surface utterance but also implicit rules often omitted from the legal documents.
It also requires following legal syllogisms, understanding the implications of related regulations, and applying them to specific case facts to deduce the final consequences.

\begin{figure}[t]
    \centering
    \includegraphics[width=8cm]{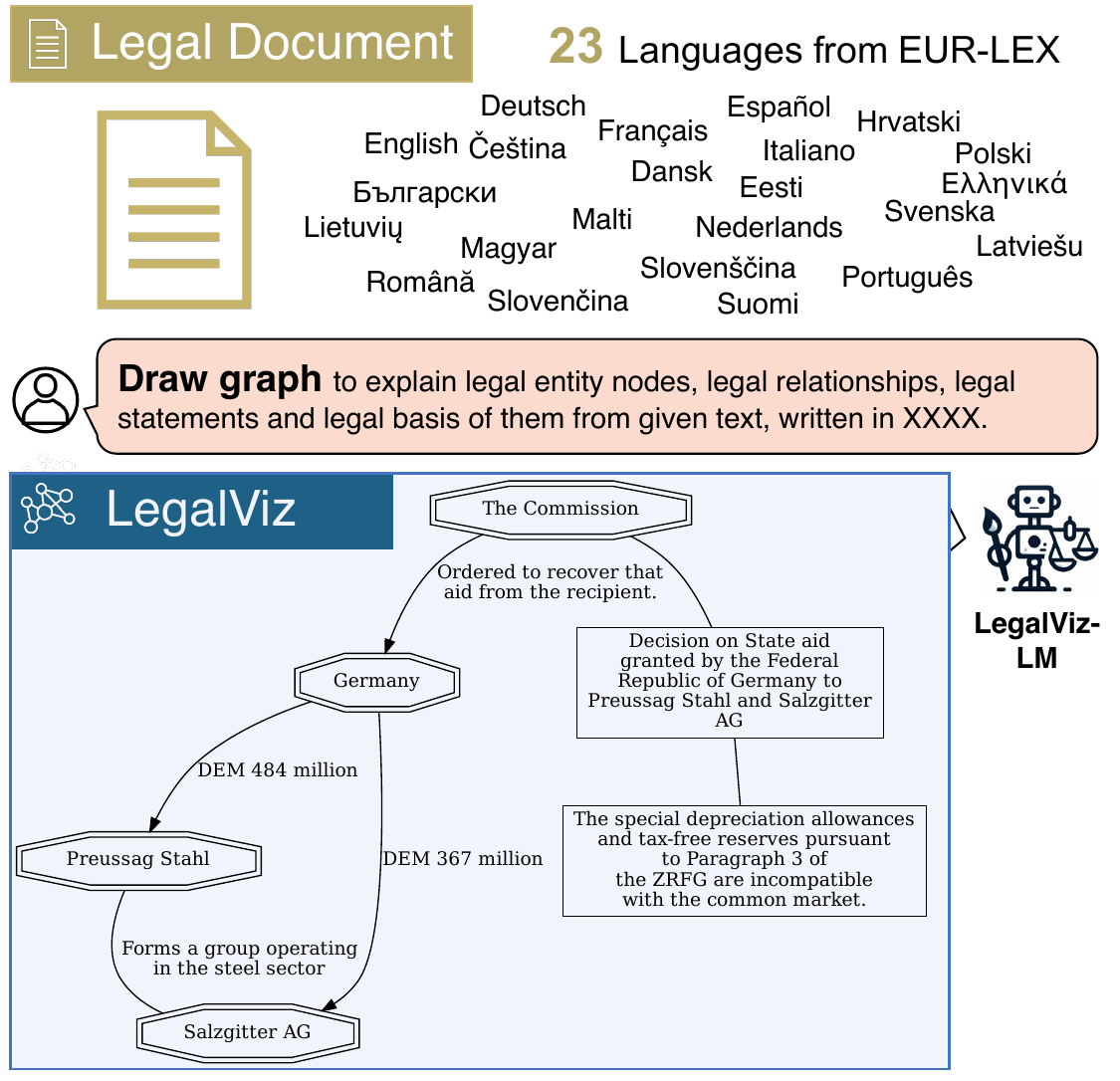}
    \caption{Model input and expected output of legal text visualization drawn by Graphviz.}
    \label{fig:teaser}
\end{figure}

At an early stage of legal NLP, there are several studies applying traditional NLP approaches for legal documents, such as named entity recognition~\citep{Angelidis2018NamedER, 10.1007/978-3-319-99722-3_32, pais-etal-2021-named, de-gibert-bonet-etal-2022-spanish}, summarization~\citep{elaraby-litman-2022-arglegalsumm, aumiller-etal-2022-eur}, text classification~\citep{chalkidis-etal-2019-extreme} and text segmentation~\citep{10.1145/3462757.3466085}.
In addition, some notable studies focus on capturing the structural legal meanings inherent in legal documents, such as learning judgment facts and results~\citep{niklaus-etal-2021-swiss}, assessing the fairness of law~\citep{chalkidis-etal-2022-fairlex}, and using the facts and attributes to predict charges~\citep{hu-etal-2018-shot}.
However, there are still numerous gaps between current legal domain status and whole automation of legal tasks by LLMs, especially judgement generation. 

The main challenges of LLM for legal applications are as follows:
(i) LLMs need to understand which legal entities are involved, the relationships between them, and the relevant legal rules.
They must also interpret the meaning of legal actions in judgments and court decisions.
If models overlook legal entities, their rights, their obligations, or key facts in interpreting the law, they fail to fulfill legal requirements, or deduce inappropriate conclusions.
(ii) LLMs must articulate why quoted rules are interpreted in their view, explaining the requirements and effects of the rules should become as they assert.
This process should adhere to the procedure of legal syllogism, requiring the recognition of potential legal entities, relationships of them, and related rules.
To address these challenges, extensive legal document resources are crucial for effectively tuning LLMs to perform well in legal domains.
However, despite the wide accessibility of plain texts of laws and court judgments on the internet,
there remains a significant lack of legal datasets with professional annotations that can enhance the capability for legal syllogism.
Moreover, LLM technology should enhance legal adaptation capabilities, supporting not only professionals but also non-experts, as everyone has legal rights and should have the opportunity to benefit from the law.

To meet these demands, we explore the novel dataset \datasetName, an automatic visualization task, generating legal diagrams that describe the legal entities, their legal relationship, related rules, and summary of the key legal facts for legal interpretation from input legal plain texts.
We introduced this visualization task with an existing diagram visualization tool of GraphViz because diagrams can succinctly elucidate complex legal relationships, allowing viewers to grasp the essentials at a glance without consulting the original article.
In fact, the visualization of legal concepts is employed in various contexts, such as textbooks for judicial examinations, university classrooms, and TV news segments.
This approach provides non-experts with easy-to-interpret visual and conceptual representations of legal materials, enhancing accessibility and understanding.
By training with our dataset, models can accurately recognize legal rules concerned in the case, identify legal entities capable of exercising rights, and understand legal transactions, and statements from professional legal documents. 
\datasetName consists of 7,010 pair professional legal documents and diagrams of DOT language code of Graphviz, with 23 languages of EUR-LEX.
Figure~\ref{fig:teaser} from the LegalViz dataset illustrates a legal diagram that explains a case where the commission required Germany to recover aid based on the common market principles, and Germany subsequently issued recovery requests.
We assume LegalViz is the first work to utilize LLMs for the visualization of legal documents.\footnote{Our dataset is available at \url{https://github.com/mizuumi/LegalViz}.}

\begin{figure*}[t]
    \centering
    \includegraphics[width=16cm]{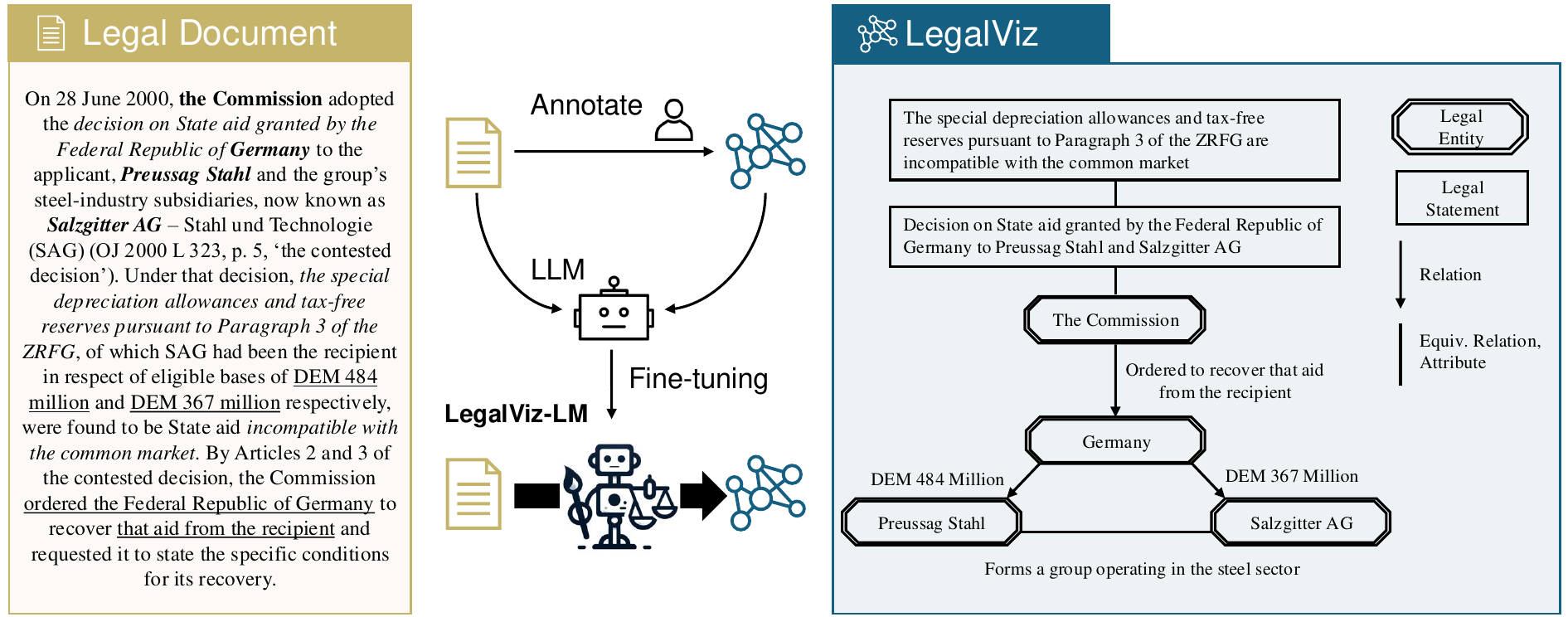}
    \caption{Legal text from EUR-LEX (left) to the resulting legal graph (right).
    }
    \label{fig:task}
\end{figure*}

\section{Related Work}

NLP applications in the legal domain are several core areas~\citep{Katz2023NaturalLP} such as information extraction, classification, summarization, judgment prediction, and resources and benchmarks.

\paragraph{Judgment prediction}
In this task, models predict the outcomes of legal cases based on given facts.
Previous studies provide judgment data from various courts of diverse countries, including decisions from the Supreme Court of the United States~\citep{10.1371/journal.pone.0174698} and the European Court of Human Rights~\citep{Medvedeva2020-MEDUML, Kaur2019ConvolutionalNN}.
Additionally, judgment prediction research has covered Switzerland~\citep{niklaus-etal-2021-swiss}, Chinna~\citep{ye-etal-2018-interpretable}, criminal law~\citep{chen-etal-2019-charge, Xiao2018CAIL2018AL}, and asylum decisions~\citep{10.1145/3086512.3086538, 10.1145/3086512.3086537}.

\paragraph{Legal resources and benchmarks} 
Datasets and benchmarks, covering a broad range of legal domains and languages, have been proposed.
These include English Tax Law~\citep{Holzenberger2020ADF}, European Legislation and the European Court of Human Rights~\citep{chalkidis-etal-2019-extreme}, Corporate and Contract Law~\citep{hendrycks2021cuad, tuggener-etal-2020-ledgar}, Supreme Court cases and US court cases~\citep{10.1145/3462757.3466088}.
The scope extends to German legal cases~\citep{icaart21}, a mixture of Korean legal text summarization, prediction and text classification~\citep{10.5555/3600270.3602628}, and refugee cases~\citep{barale-etal-2023-asylex}. 
Additionally, multilingual and multi-legal domain datasets have been developed, such as a multilingual corpus of English, German, Italian, Polish ~\citep{drawzeski-etal-2021-corpus},
and LEXGLUE~\citep{chalkidis-etal-2022-lexglue} which covers six predictive tasks over five datasets made of English from the US, EU, and Council of Europe.
Furthermore, Lexfiles~\citep{chalkidis-etal-2023-lexfiles} offers a comprehensive dataset of comprised of US, UK, Canada, India, European Court of Human Rights, and Lextreme~\citep{niklaus-etal-2023-lextreme}, which covers wide-range of tasks and countries among EU nations.
However, none of these datasets are designed to support the visualization of legal documents for non-experts.
In contrast, \datasetName offers legal specific annotations in 23 multilingual legal documents, specifically tailored for visualization.
These annotations cover legal entities, their relationship, related rule, related facts of legal texts, thereby enhancing the clarity and interpretation of legal documents for judicial judgments.

\if[]
Our \datasetName contributes to identification of legal entity, their relationship, related rule, related facts of legal texts, providing 
Once those four elements are identified from dispute backgrounds, judges can generate legal decisions using law identified by model and applying entities and facts to the legal requirements.
\fi

\paragraph{Text to graph generation}
Following the iconic successions of the GPT models, LLMs can generate not only contextual texts and program codes~\citep{shi-etal-2022-natural, christopoulou-etal-2024-text} but also visualization codes~\citep{SparksofAGI}, such as creation of scientific vector graphics with TiKZ~\cite{automatikz}
and diagram generation with refinements and diffusion process~\cite{Zala2023DiagrammerGPT}.
Text-to-code generation studies are predominantly focused on mainstream programming languages like Python and shell scripts, and are typically examined with English text~\citep{shi-etal-2022-natural, christopoulou-etal-2024-text}.
Both text-to-graph generation and graph-to-text generation studies are often conducted for clarifying paragraph structure and summarizing critical issues and relationship between words~\citep{koncel-kedziorski-etal-2019-text, jin-etal-2020-genwiki} of the input plain texts mainly in English.
These text-to-graph approaches are suitable for free drawing based on text instructions, but they sacrifice the visualization of logical relationships within the visualized content. 
In comparison, our graph generation approach utilizes the DOT language of Graphviz, enabling models to focus specifically on visualizing the logical relationships within the content.

\section{Building LegalViz Dataset}

\subsection{Legal Visualization}
\label{sec:legalviz}
The aim of legal visualization tasks is to generate an interpretable graph that clarifies the legal relationships embedded within the input legal texts. 
The constructed graph comprises legal entities and/or rules as nodes, connected by edges representing legal transactions and/or significant facts relevant for judicial determination. 
To effectively visualize these legal relationships, we utilize the DOT language of Graphviz, a widely adopted open-source tool for graph visualization. 
Figure~\ref{fig:task} presents an overview of our proposed task, showcasing both the expected input and output.

\paragraph{Legal entity}
Legal entities are applicants and respondents of judgment, courts, creditors, debtors, criminal suspects, or companies and employees.
Legal entities are represented in \textbf{double octagons}.
In contrast to grammatical general nouns, proper nouns, or objects, we concentrate on persons or organizations capable of exercising legal rights and engaging in transactions.

\paragraph{Legal relationship \& transactions}
Legal relationships encompass various form of relationships between legal entities, including the exercise of legal rights from one to another, legally significant transactions, the interrelations between legal statements made by entities and the underlying norms that support them, and relationships defined under law such as employment, contractual agreements, marriage, and family relationships.
Legal transactions are specific types of relations among legal entities, such as purchases, notifications and any actions exercising rights.
Both legal relationships and transactions are represented in the directed or undirected \textbf{edges} with various styles between legal entities.
Some edges have textual relation labels.

\paragraph{Legal source}
Legal sources are rules applied or referred by court and support legal statements. 
Here we concentrate on legal sources explicitly written in the legal document.
They include constitutions, statutes, ordinances, and case laws.
These extracted rules are represented in \textbf{trapeziums}.
Each trapezium of the legal source is connected to a node interpreting rules supported by the legal source via undirected edges.

\paragraph{Legal statement}
Legal statements are detailed explanations of transactions and factual descriptions of the case notable for the final judgment to summarize.
Adding these summaries to diagrams help non-experts grasp the facts important for final judgments at a glance.
Legal statements are represented in \textbf{squares}, connected to a node by an edge.

\begin{table}[t]
\centering
  \footnotesize
\begin{tabular}{lrrr}
\toprule
\textbf{Split} & \textbf{\# Instances} & \textbf{\# Nodes} & \textbf{\# Relations}\\
\midrule
Train & 4,710 & 12,624 & 16,367 \\
Validation & 1,150 & 3,404 & 2,717 \\
Test & 1,150 & 3,128 & 3,589 \\
\midrule
Total & 7,010 & 19,156 & 22,673 \\
\bottomrule
\end{tabular}
\caption{Dataset splits.}
\label{table:dataset_split}
\vspace{-1em}
\end{table}

\begin{table}[t]
\centering
  \footnotesize
\begin{tabular}{lccccc}
\toprule
\textbf{Lang.} & \textbf{ISO} & \textbf{\# Ins.} & $L_{\mathrm{word}}$ & $L_{\mathrm{char}}$ & $L_{\mathrm{code}}$ \\
\midrule
All & - & 7,010  &  109.0  &  644.2  &  759.8 \\
\midrule
Bulgarian & BG & 290  &  113.4  &  625.5  &  759.8 \\
Spanish & ES & 307  &  133.7  &  693.4  &  708.4 \\
Czech & CS &  307  &  102.8  &  582.9  &  832.5 \\
Danish & DA & 307  &  110.9  &  640.7  &  766.8 \\
German & DE & 312  &  108.9  &  683.0  &  759.1 \\
Estonian & ET & 307  &  83.9  &  588.8  &  809.4 \\
Greek & EL &  307  &  121.4  &  698.6  &  779.2 \\
English & EN & 312  &  122.6  &  629.2  &  623.0 \\
French & FR &  312  &  128.6  &  674.8  &  766.9 \\
Croatian & HR & 263  &  103.2  &  577.7  &  718.7 \\
Italian & IT &  312  &  123.3  &  705.1  &  788.7 \\
Latvian & LV & 307  &  94.4  &  598.8  &  725.7 \\
Lithuanian & LT & 307  &  94.6  &  609.4  &  749.8 \\
Hungarian & HU & 307  &  97.4  &  670.2  &  809.7 \\
Maltese & MT & 305  &  100.4  &  706.3  &  777.7 \\
Dutch & NL & 312  &  122.0  &  687.0  &  784.7 \\
Polish & PL & 307  &  106.7  &  655.0  &  759.2 \\
Portuguese & PT &  307  &  125.2  &  653.1  &  778.0 \\
Romanian & RO & 290  &  118.3  &  672.0  &  791.8 \\
Slovak & SK &  308  &  101.0  &  585.7  &  727.9 \\
Slovenian & SL & 308  &  106.6  &  580.0  &  730.5 \\
Finnish & FI & 308  &  78.5  &  649.6  &  808.7 \\
Swedish & SV & 308  &  108.9  &  639.1  &  748.2 \\
\bottomrule
\end{tabular}
\caption{
Dataset statistics.
$L_{\mathrm{word}}$ and $L_{\mathrm{char}}$  are length of legal text. $L_{\mathrm{code}}$ is character length of Graphviz code. 
}
\label{table:dataset_statistics_1}
\vspace{-1em}
\end{table}

\subsection{Dataset Creation}
\paragraph{Collection of legal document}
To construct the legal graph dataset, we collected legal documents in the following manner:
(i) We sourced legal documents from the EUR-LEX website\footnote{\url{https://eur-lex.europa.eu}}, which provides public access to judgments, orders, and rules of EU countries in official EU languages.
We specifically selected judgments from the years 2006 to 2019, available in translations across 23 languages, to capture the latest legal trends. 
(ii) We extracted the factual sections of the judgments that contain legal facts to be expressed in the graph. 
(iii) Finally, we gathered the corresponding sections of legal documents in 23 languages to ensure consistency across translations.

\paragraph{Graphviz annotation}
We have manually annotated Graphviz code visualization from the legal documents by an annotator with expertise in the legal domain.
The process involved several steps:
(i) We broke down long judgment cases into short paragraphs so that DOT language can draw diagrams in units easily understandable at a glance.
(ii) We extracted the legal entities and rules as nodes of the diagram, legal transactions as edge relations within the diagram, and the summary of statements and explanations as squared nodes.
(iii) We created a Graphviz diagram to represent the extracted relations, using variations in node shape and relations, as defined by the rules in Appendix~\ref{sec:graph_formalism}.

\paragraph{Translation of Graphviz annotation}
To cover the European Union's official languages present at the time the judgment was written, we translated our English annotation to other languages as follows:
(i) We first used GPT-4 to extract the legal words and sentences from the provided English judgements, aiming to save as many terms as possible from the EU's officially translated variations of judgments.
(ii) If GPT-4 fails the extraction task, we then apply GPT-4 translations from English to other languages.
(iii) We manually checked the previously translated sentences and retranslated them using DeepL and the Azure GPT API if any translation errors were detected.
The prompts used in the translation process are described in Appendix~\ref{sec:appendix}.

\paragraph{Dataset statistics}
Table~\ref{table:dataset_split} shows the statistics of the \datasetName dataset.
We build a total of 7,010 pairs of legal texts and graphs, encompassing 23 language variations and more than 300 unique legal texts.
The constructed legal graph consists of 19,159 nodes and 22,673 relations.
We also summarize the average word length, number of characters in legal sentences, and character length of Graphviz code for each language in Table~\ref{table:dataset_statistics_1}.

\section{Evaluation}
\label{sec:eval}
We compare the reference and hypothesis graphs to assess the quality of the generated.
One straightforward way to achieve this is to directly compare two images visualized by GraphViz.
However, this approach clearly ignores the textual structure of the legal documents.
One other approach is directly comparing the DOT language codes in textual manner, ignoring numerous minor differences of the visualization codes that can result in different graphs.
Therefore we propose an approach to first compose graphs, align the graph components, and then compare each component of graphs using textual metrics as described in this section.

\input{tables/main_result}

\subsection{Similarity of two graphs with texts}
Formally, let $\mathcal{G}_r$ and $\mathcal{G}_h$ be the reference and hypothesis graphs.
Each graph is composed of a set of edges $E$ and nodes $V$.
When an edge $e\in E$ connects a starting node $v_s$ and an end node $v_e$, it is represented by a tuple $e=[v_s,v_e,l]$, where $l$ is a label of this edge.
Nodes always include non-empty texts, while edge-label texts can be blank for edges without labels.

\paragraph{DOT code validation}
First, we examine whether the generated Graphviz code forms a valid graph $\mathcal{G}_h$ in terms of the DOT language. This is done by simply processing with the pydot library\footnote{https://github.com/pydot/pydot}.

\paragraph{Nodes alignment by bipartite matching}
Second, we extract nodes $V_h$ from $\mathcal{G}_h$ and align them with nodes from the reference graph: $V_r$ from $\mathcal{G}_r$ using the similarity of the texts in nodes.
For this node alignment, we apply the bipartite matching problem for the two sets of nodes $V_h$ and $V_r$, using the similarity function $\mathrm{sim}(v_r,v_h)$ between two texts in the reference $v_r \in V_r$  and hypothesis nodes $v_h \in V_h$ with a textual similarity metric.
Here we use BERTScore~\cite{bertscore} for the similarity metric because of its robustness and high human correlation.
Given the textual similarity scores between all reference and hypothesis nodes, we apply a bipartite matching solver in NetworkX\footnote{https://networkx.org/} for nodes and obtain the nodes alignment between the reference and hypothesis graphs that are used for later evaluation.

\paragraph{Graph, node \& edge evaluation}
After we determined the node alignment, we performed the comparison of the two graphs based on the nodes, edges and their labels.
We introduce the following three metrics with different depth:
\texttt{Graph}, \texttt{Graph\&Node} and \texttt{Graph\&Node\&Edge}.

\texttt{Graph} is an F1 metrics of the matched edges after the node alignment by bipartite matching. 
This metric is for the similarity measurement of the entire graph structure, ignoring the textual differences of nodes and edges after the alignment.
Let the node set $V_r$ and edge set $E_r$ composes reference graph $G_r$, and the node set $V_h$ and edge set $E_h$ composes hypothesis (generated) graph $G_h$.
Using a node alignment function $a(\cdot): V_h \to \{V_r,\phi \}$ from the generated graph nodes $V_h$ to the reference graph nodes $V_r$ and Kronecker delta $\delta_{\mu \nu} = 1$ iif $\mu=\nu$ otherwise $\delta_{\mu \nu} = 0$, which represents the agreement of the nodes here,
we compute the agreement score of the nodes as
\begin{align}
    \nonumber f_{\mathrm{Graph}}(e_h,e_r)&=\delta_{a(v_{s,h}) v_{s,r}} \delta_{a(v_{e,h}) v_{e,r}}
\end{align}.
$v_{s,h}$ and $v_{s,r}$ are the start nodes of each edge from the hypothesis (generated) and reference graphs, while $v_{e,h}$ and $v_{e,r}$ are the end nodes of it, respectively. 
The generated nodes can be aligned to $\phi$ when they are not aligned to any reference nodes: $v_h \xrightarrow{a(\cdot)} \phi$. Here $\phi$ is a null node, and we assume for any nodes $\nu$, $\delta_{\phi\nu}=0$. From this binary function $f_{\mathrm{Graph}}(e_h,e_r)$, we can compute TP, FP and FN by
\begin{align}
\nonumber 
    \mathrm{TP} &= \sum_{e_h \in E_h,e_r \in E_r} f_{\mathrm{Graph}}(e_h,e_r) \\
    \nonumber 
    \mathrm{FP} &= |E_h|-\mathrm{TP}, \ \ \ \ \ ~~ 
    \mathrm{FN} = |E_r|-\mathrm{TP}
\end{align}
and obtained F1 value from these for \texttt{Graph}.

\texttt{Graph\&Node} is an F1-based metric where BERTScore penalize the dissimilar texts of the two aligned nodes pairs $\{v_{s,h}, v_{s,r} \}$ and $\{v_{e,h}, v_{e,r} \}$ for each edge. TP is calculated as
\begin{align}
\nonumber 
    \mathrm{TP} = \sum_{e_h \in E_h,e_r \in E_r} &f_{\mathrm{Graph}}(e_h,e_r) \ \cdot \\ 
    \nonumber &\mathrm{sim}(v_{s,r},v_{s,h})\mathrm{sim}(v_{e,r},v_{e,h})
\end{align}
while FP and FN are calculated from TP.
Because of the products of the start and end node similarities, the \texttt{Graph\&Node} metric is sensitive to the difference of node texts compared with the \texttt{Graph} metric.

Similarly, \texttt{Graph\&Node\&Edge} is an F1-based metric that considers node and edge text similarity in terms of the BERTScore as following
\begin{align}
\nonumber 
    \mathrm{TP} = &\sum_{e_h \in E_h,e_r \in E_r} f_{\mathrm{Graph}}(e_h,e_r) \ \cdot \\
    \nonumber &~~~~~~\mathrm{sim}(v_{s,r},v_{s,h})\mathrm{sim}(v_{e,r},v_{e,h}) \mathrm{sim}(l_r,l_h)
\end{align}
by penalizing dissimilar texts of the edge texts.

\subsection{Evaluation of Legal Content}
\label{sec:legalvizeval}
We also introduce the evaluation of legal contents in our visualizations. As described in Sec.~\ref{sec:legalviz}, the legal contents in LegalViz are associated with specific diagrams in GraphViz. For diagrams of legal entities (\textbf{double octagon}), Legal source (\textbf{trapeziums}), Legal statement (\textbf{squares}), we extract these nodes from the reference ($v_r$) and hypothesis ($v_h$) graphs and check whether they are properly aligned in the alignment in the previous section. Then we measure the similarity of the node texts with BERTScore for successfully aligned nodes and compute micro averaged F1 score as following $\mathrm{TP}$, $\mathrm{FP}$, $\mathrm{FN}$:
\begin{align}
    \mathrm{TP} &= \sum_{v_h \in V_h, v_r \in V_r} \delta_{v_r,v_h} \mathrm{sim}(v_r,v_h) \nonumber \\ 
    \mathrm{FP} &= |\{v_h\}|-\mathrm{TP} \nonumber \\
    \mathrm{FN} &= |\{v_r\}|-\mathrm{TP}
    \nonumber
\end{align}
where $\delta_{v_r,v_h}=1$ iff $v_r$ and $v_h$ is aligned, and otherwise 0.
For legal relations \& transactions (\textbf{edges}), we extract the aligned edge label texts and compute F1 score from the similarity of labels.

\input{tables/multi_lang_parts}

\section{Experiments}
We evaluate the ability to visualize graphs from legal sentences with \datasetName.
Overall, our finetuned models overperformed fewshot GPT models.

\subsection{Experimental settings}
We conduct the experiments of legal visualization in the manner of the DOT language code generation with the publicly available Llama and Gemma family models and OpenAI GPT APIs via Microsoft Azure. We use the GPT-3.5-Turbo (\texttt{1106}), GPT-4 (\texttt{0613}) and GPT-4o (\texttt{2024-05-13}) models.
For Llama family models, we experimented with the models specialized for code generation of CodeLlama~\citep{Rozire2023CodeLO} and the recently released Llama 3.1 \& 3.2 models~\citep{dubey2024llama3herdmodels} and Gemma 2-9B~\citep{Riviere2024Gemma2I} models.
Experimental settings are two holds: few-shot generation and finetuning of the publicly available models.
In few-shot experiments, we notice not only the GPT models but only publicly available models are capable of producing valid DOT language codes without finetuning to some extent.
We follow the supervised finetuning of Hugging Face with the detailed finetuning parameters in Appendix~\ref{app:param}.
In evaluation, we generate ten different Graphviz codes with each model and examine their quality in evaluation methods of graph and legal contents described in Sec.~\ref{sec:eval}.

\subsection{Result}
\paragraph{Graph-based Evaluation}

The Graph-based Evaluation section of Table~\ref{table:generation_performance} presents the experimental results of each model evaluated by \texttt{Graph}, \texttt{Graph\&Node}, and \texttt{Graph\&Node\&Edge} metrics explained in Section~\ref{sec:eval}.
Among three evaluation metrics, \texttt{Graph\&Node\&Edge} is the most difficult because all three graph elements must be correct as shown in the evaluation.
Most importantly, our finetuned models outperformed few-shot counterparts and even GPT models, which are assumed to be larger than the CodeLlama-13B models, suggesting the effectiveness of our dataset for finetuning. 
Also, finetuned Gemma-2-9B took the highest scores on \texttt{Graph} and \texttt{Graph\&Node}, and \texttt{Graph\&Node\&Edge}. Surprisingly, Gemma-2-9B performed worse than Gemma-2-9B-it before finetuning, suggesting the effectiveness of finetuning with LegalViz.

\begin{figure*}[t]
    \centering
    \includegraphics[width=15.5cm]{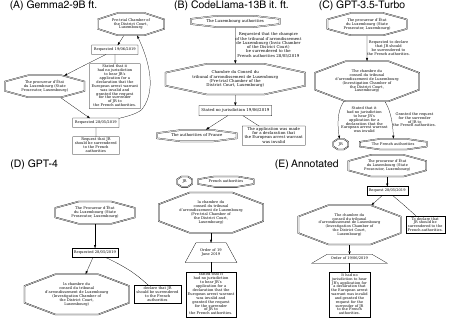}
    \caption{Qualitative analysis of diagrams by Graphviz code. Figures are generated by the finetuned modesl of Gemma2-9B, CodeLlama-13B-Instruct, few shot models of GPT-3.5-Turbo, GPT-4 and an annotated diagram.}
    \label{fig:qualitative}
\end{figure*}

\paragraph{Valid DOT code ratio}
The Valid Graph section of Table~\ref{table:generation_performance} presents the success rate of forming a valid DOT language code without code syntax errors.
In the first generation trial, GPT-4 is the most accurate to generate valid DOT language codes among all models in both few-shot and finetuned settings and the second best is Gemma2-9B finetuned model.
When we let models generate ten variations, several finetuned models (Llama-3.1-8B, Gemma2-9B, Gemma2-9B-it) are able to generate valid DOT code in 100.0 percents of the test set.
Comparing the publicly available few-shot and finetuned models, finetuning with our dataset strongly improves the valid graph creation of all models.

\paragraph{Legal Content Evaluation}
The Legal Content section of Table~\ref{table:generation_performance} presents the legal aspect-wise evaluation as described in Sec.~\ref{sec:legalvizeval}.
By nature of the legal entities, \texttt{Entity} can be extracted from input sentences in many conditions, achieving high scores in the table.
However, the other three aspects aren't easily extracted from the input legal text.
\texttt{Statement} includes the text generation for legal facts and tends to be lower scores than others. 
This is because legal statements appear in texts without some remarkable keywords, compared to legal \texttt{Source}, which is often mentioned in texts with terminology such as ``Law'' and ``Act'' and legal \texttt{Relations \& Transactions}, which is found in texts with terminologies such as ``contract,'' ``issue'' with some warrants and orders, ``notification'' with notable as a legal act.
\texttt{Statement} acts for summarizing notable facts related to rule and its interpretation in question, to describe the detail of other nodes especially legal relations and transactions, and to explain the facts applicable to legal requirements.
Finetuned \texttt{Gemma2-9B} achieved highest score in all four aspects of the legal content evaluation.
The scores in \texttt{Statement} are improved by approximately three times compared to the few-shot scores across all models, suggesting the effectiveness of finetuning with our dataset.

\paragraph{Scores by languages}
Table~\ref{table:scores_by_language} presents the results in legal contents by all 23 languages in EUR-LEX.
We present the best performing model of Gemma2-9B in finetuned and fewshot conditions. We also present the averaged results of 10 models in Table~\ref{table:generation_performance} without GPTs to highlight the performance difference before and after finetuning with LegalViz across languages, while minimizing the influence of individual model characteristics.
Among these languages, models perform relatively weakly in languages that have relatively fewer resources~\citep{chalkidis-etal-2021-multieurlex}, such as Maltese, Latvian, Lithuanian, 
 Hungarian. For languages that have relatively more resources such as English and French, models tend to have higher scores than others.
From a linguistic point of view, Hungarian and Finnish, belonging to the same Uralic language group, have low scores in each model. This may reflect their linguistic difference from other languages.
Similarly, for the Romance language group, e.g., Romanian, French, Spanish, Italian, and Portuguese, models have moderate performances, seemingly better than those of the Uralic language group and languages that also have fewer resources than those of English and French.
Among four legal aspects, the source and statement parts include the summarization task of the legal document for visualization. They are considerably difficult parts in the graphs and the performance in some languages becomes 0 without finetuning.
It is also notable that finetuning contributes the performance in all aspects in all of these languages.

\subsection{Qualitative Analysis}
\label{sec:qualitative}
We conduct a qualitative analysis using the best performing model of the finetuned Gemma2-9B and CodeLlama-13B Instruct models, along with the few-shot GPT-3.5-Turbo and GPT-4.
Figure~\ref{fig:qualitative} presents the graphs generated from English legal documents along with the annotated graph.
This legal document used for the model input is on the Appendix~\ref{app:inuput}.
This is a part of a criminal procedural case in which the prosecutor requested the court to declare securing custody where the prosecutor and the court are legal entities.
Square nodes in annotated data are describing intention of request and consequence of the request written in order.

For fewshot models of GPT-3.5-Turbo and GPT-4, GPT-3.5-Turbo wrongly recognize that all nodes are legal entities and illustrate them in double octagon.
Its description of the legal relationship between the court and the authority is also incorrect because the court didn't take the legal action directly in this article. 
The GPT-4 model assumes that ``JR'' and ``French authorities'' are legal entities but fail to illustrate relationships between those and other entities as these entities lack connections to other graph parts. 
Also, GPT-4 extracted two different court names from given text as different entities. However they are indeed different divisions of the same court and fails to summarize the relationship between them.
Gemma2-9B ft. successfully extracts the detailed description of court's request, while it extracted incorrect entity as double octagon shape and it couldn't extract the court order and its description.
CodeLlama-13B-Instruct ft. model successfully forms the graph structure but the square node mentioned European arrest warrant in the right bottom is inconsistent with input text.

\section{Conclusion}
We have proposed \datasetName, the first manually annotated dataset to visualize legal text with DOT language Graphviz and introduced a novel evaluation method taking into account both diagram visualization quality and sentences of not only graph nodes and relations but also legal contents.
We empirically confirmed the effectiveness of our dataset with wide-range of experiments including comparisons of few-shot and finetuning models and demonstrated trained models outperform the closed models of GPTs in all evaluation metrics.

\section*{Limitation}
\datasetName contains the same number of instances in
23 languages of EUR-LEX. However, this doesn't mean that the models with fintuned or few-shot have the same ability to treat all 23 languages equally. Especially models face difficulties in fewer language resources as we experimented.
We cannot offer any warranty for using our dataset and models for real usages such as legal advice. We also consider that our dataset should be used with appropriate supervision by experts. This can be a \textit{potential risk} when our dataset is misused. We assume that results of automatic visualizations by models are still different from the annotated visualizations in most cases, suggesting the current limitation of the generation.

\section*{Ethic Statements}
The annotation material of this dataset is publicly available EU legal materials including judgments and orders, which do not include personal or sensitive information, with the exception of trivial information presented by consent, e.g., the names of the active presidents of the European Parliament, European Council, or other official administration bodies. The copyright for the editorial content of this website, the summaries of EU legislation, and the consolidated texts, which are owned by the EU, is licensed under the Creative Commons Attribution 4.0 International license.\footnote{https://eur-lex.europa.eu/content/legal-notice/legal-notice.html}

\section*{Acknowledgment}
This work was supported by JSPS Grant-in-Aid for Young Scientists (JP22K17983), 
 JSPS Fostering Joint International Research (A) (JP22KK0184)
 and JST CRONOS JPMJCS24K6.

\bibliography{anthology,custom}

\onecolumn
\appendix

\section{Qualitative analysis input}
\label{app:inuput}
The English legal text used the qualitative analysis in Section~\ref{sec:qualitative} is the following:

\noindent\fbox{%
    \parbox{\textwidth}{%
On 28 May 2019, the procureur d’État du Luxembourg (State Prosecutor, Luxembourg) requested that the chambre du conseil du tribunal d’arrondissement de Luxembourg (Investigation Chamber of the District Court, Luxembourg) declare that JR should be surrendered to the French authorities. By order of 19 June 2019, the chambre du conseil du tribunal d’arrondissement de Luxembourg (Pre-trial Chamber of the District Court, Luxembourg) stated that it had no jurisdiction to hear JR’s application for a declaration that the European arrest warrant was invalid and granted the request for the surrender of JR to the French authorities.
    }%
}

\section{Detailed experimental settings}
\label{app:param}

For training of LLMs, we follow the default setting of Hugging Face supervised finetuning of the trl\footnote{https://github.com/huggingface/trl} library for the optimizers and schedulers. We use the mini-batch size of 32. We use the max token length of 4096 for training as we notice some languages, e.g., Greek, require longer tokens than other languages depending on Llama tokenizers. In finetuning, we use FP32 precision and all trainable parameters are updated. All Llama-family experiments are done on a single node with four NVIDIA A100 GPUs.

\input{tables/multi_lang_parts_appendix}
\input{tables/multi_lang_gne_appendix}
\input{tables/main_result_appendix}

\section{Results of the Validation split}
Table~\ref{table:main_result_appendix} shows the results of the validation split of the main performance table of Table~\ref{table:generation_performance}.

\section{Additional Multilingual Experiments}
We experimented several models listed in the tables below and selected models with great scores are discussed in main paper.

\paragraph{Multilingual results of all models}
We conducted experiments using the following models, namely, Llama3.1-8B, Llama3.1-8B-Instruct, Llama3.2-3B, Llama3.2-3B-Instruct, CodeLlama-7B, CodeLlama-7B-Instruct, CodeLlama-13B, CodeLlama-13B-Instruct, Gemma2-9B, and Gemma2-9B-it.
Firstly, the results evaluated with legal content evaluation point of view are shown in Table~\ref{table:scores_by_language_all_fewshot} and Table~\ref{table:scores_by_language_all_finetuned}.
Table~\ref{table:scores_by_language_all_fewshot} shows few-shot Legal Content evaluation of all models conducted and Table~\ref{table:scores_by_language_all_finetuned} shows finetuning results of Legal Content evaluation.  
Secondly, the results evaluated with graph-based point of view are summarized in Table~\ref{table:scores_by_language_fewshot} and Table~\ref{table:scores_by_language_finetuned}.
Table~\ref{table:scores_by_language_fewshot} shows few-shot results of graph-based evaluation and Table~\ref{table:scores_by_language_finetuned} represents.

For comparison of each language's score, we calculated the average of all few-shot models and the average of few-shot models despite GPT models in Table~\ref{table:scores_by_language_all_fewshot} and Table~\ref{table:scores_by_language_fewshot}.
In the same way, the average score of all finetuned models is calculated in Table~\ref{table:scores_by_language_all_finetuned} and Table~\ref{table:scores_by_language_finetuned}.

\section{Applications of traditional NLP tasks for legal domain}
\paragraph{Legal information extraction}
Named Entity Recognition (NER) is a fundamental information extraction task that has been developed for several languages, including 
Greek~\citep{Angelidis2018NamedER}, Brazilian~\citep{10.1007/978-3-319-99722-3_32}, Romanian~\citep{pais-etal-2021-named}, and Spanish~\citep{de-gibert-bonet-etal-2022-spanish}.
Those NER approaches extract mainly the same objects as those in non-legal domains, while some efforts try to extract legal entities from court documents~\citep{Chapter11LexNLPNaturallanguageprocessingandinformationextractionforlegalandregulatorytexts}.
Once NER identified entities, Relation Extraction in the legal domain~\citep{chalkidis-etal-2021-paragraph} takes this information further by identifying and classifying the relationships between these entities, such as facts and allegedly violated articles, specific articles and paragraphs, and case references, as well as relevant facts and allegations.

\paragraph{Legal classification} 
The classification task of legal texts has been proposed with a focus on practical applications.
For example, to enhance the interpretation of complex legal information, multi-label classification of legal texts assigns multiple conceptual class labels to words appearing in legal sentences \citep{chalkidis-etal-2019-extreme}.
Notably, FairLex~\citep{chalkidis-etal-2022-fairlex} aims to ensure the fair application of the law by classifying attributes such as age, gender, region, and state.

\paragraph{Legal summarization} 
As a more complex and application-oriented task, legal summarization is also prominent in the field, which aims to generate a summary of legal sentences.
Existing summarization studies address Canadian legal cases~\citep{elaraby-litman-2022-arglegalsumm} and EU legislations~\citep{aumiller-etal-2022-eur}.

\section{Legal Diagram Formalism}
\label{sec:graph_formalism}

Here we define several rules to express legal relations within the DOT language grammar.

\paragraph{Graph node rules}
Legal entities are represented by nodes (vertices) in DOT languages with the shape of double octagons except legally deceased persons who are presented in the shape of ellipses.
Legal norms that are effective in the present case are represented by graph nodes with  trapezium shapes.

\paragraph{Graph edge rules}
Legal transactions and the explanatory relationships between legal entities are represented by directed edges.
The family or marital relationships established under civil law are represented by an undirected bold edge.
The legal rights that cannot be exercised are represented by dashed edges. 
Dotted edges denote relationships of the legal succession between legal entities.

To illustrate the equivalent relationship between diagram nodes, undirected edges are used to connect entities and their status explanations, rules and statements, legal transactions, and their explanations.

\begin{figure}[t]
    \centering
    \includegraphics[width=7.5cm]{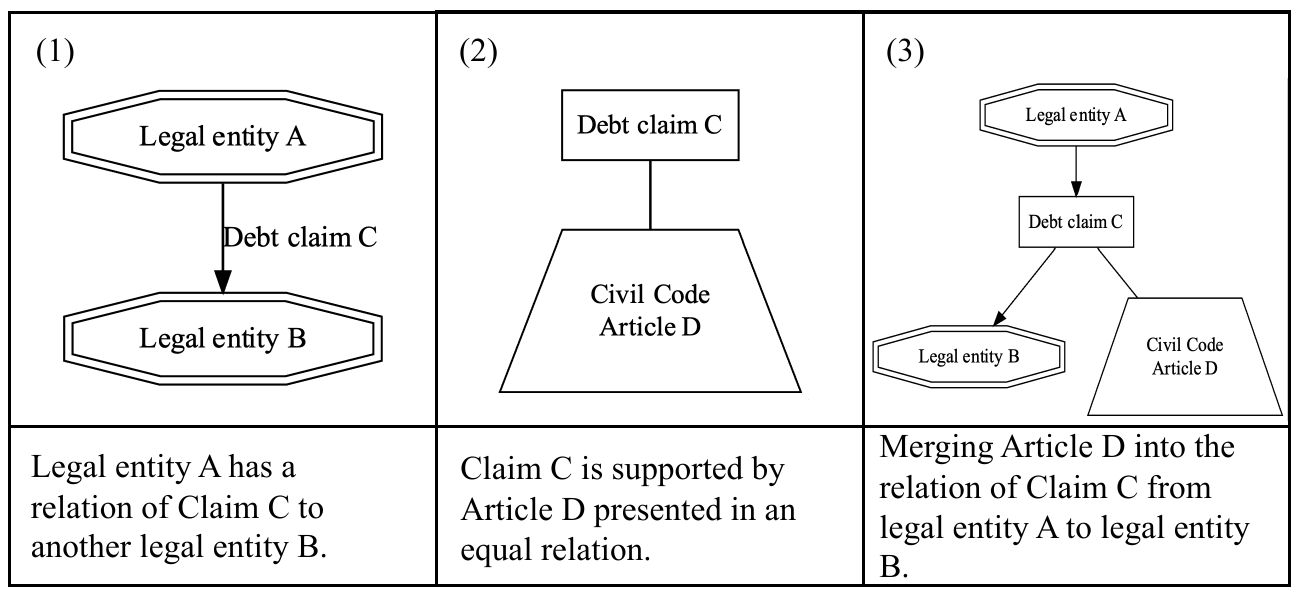}
    \caption{
    Annotation rule when adding explanation to graph relations.
    }
    \label{fig:annotation}
\end{figure}

We also note that legal relations can also be represented by graph nodes when legal relations have some relations with other entities.
Figure~\ref{fig:annotation} explains how to draw graphs when additional description is required for graph relations.
In Graphviz, we cannot draw lines directly to the graph relations.
Hence we change graph labels to nodes and connect to other nodes for adding explanation.
Further details of the DOT language grammar for representations of legal entity relations and an actual dataset example are provided in Appendix~\ref{app:anno rule}~\&~\ref{app:dataset}.

\section{Graphviz annotation}
\label{app:anno rule}
The following is an example of the Graphviz code annotation rules.

\lstset{
    frame=single,
    numbers=left,
    tabsize=2
}
\begin{lstlisting}
[shape=doubleoctagon]: Entities which are capable to act as legal entity.
[shape=trapezium]: Any kinds of rules which are legally effective, applied to the present case or supporting legal statements.
[style=dotted]: Relationship of succession between 2 entities.
[dir=none]: Equivalent relationship, agreements, or connecting detailed explanation of other nodes.
[dir=none, style=bold]: Marital relationships or family relationships which have been established under civil law.
[style=dashed]: Expressing a legal right that cannot be exercised or not existed.
[shape=ellipse]: Expressing a person who is legally deceased.
\end{lstlisting}

\section{Qualitative Analysis}
\begin{figure}[t]
    \centering
    \includegraphics[width=15cm]{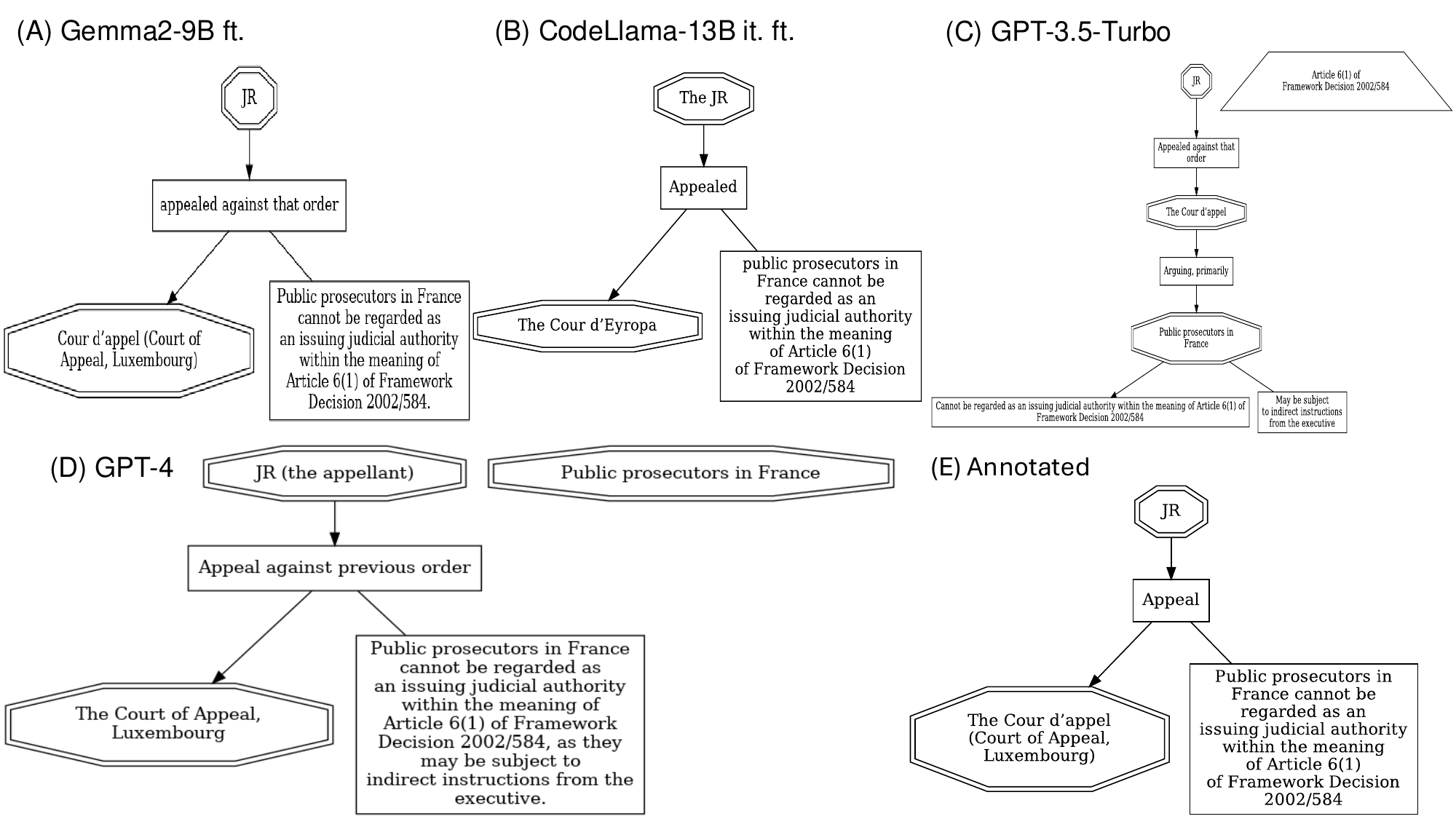}
    \caption{
    Additional qualitative analysis.
    }
    \label{fig:qualitative_appendix}
\end{figure}

Figure~\ref{fig:qualitative_appendix} shows an additional case of qualitative analysis. 
Here, GPT-3.5-Turbo and GPT-4 failed to generate correct relational graph since some nodes (``Public prosecutors in France'' and ``Article 6(1) of Framework Decision 2002/584'') are not connected to other nodes lacking understandings of legal relations.
On the other hand, Gemma2-9B ft. and CodeLlama13B-Instruct ft. models fine-tuned with LegalViz output almost the same diagram as annotated data, with the same legal entities (``JR'' and ``The Court of Appeal in Luxembourg'') and correctly extracted reason of appeal.

\section{Prompt}
\label{sec:appendix}
The prompt for LLMs used in training, generation and dataset creation is presented in Table~\ref{tab:app-prompt}.

\input{tables/prompt_table}

\section{Train dataset examples}
\label{app:dataset}
\subsection*{Dataset Example}
\begin{center}
\begin{tabular}{c}
{\scriptsize
\begin{lstlisting}[breakatwhitespace=false]
{'ID': '45',
 'category': 'EU law',
 'diagram_number': '7',
 'case_name': 'Case T-207/02: Nicoletta Falcone v Commission of the\nEuropean Communities',
 'case_number': 'C2005/006/64',
 'document_url': 'https://eur-lex.europa.eu/legal-content/EN/TXT/PDF/?uri=CELEX:C2005/006/64&qid=1713891140330',
 'year': '2004',
 'text': 'In Case T-207/02: Nicoletta Falcone, a candidate in Competition COM/A/10/01, represented by M. Condinanzi, against Commission of the European Communities (Agent: J. Currall, assisted by A. Dal Ferro, with an address for service in Luxembourg) - application for annulment of the decision of 2 May 2002 of the selection board in Competition COM/A/10/01 to exclude the applicant from the written tests on the ground that she did not obtain sufficient marks to be included among the 400 best candidates - the Court of First Instance (Second Chamber), composed of J. Pirrung, President, A.W.H. Meij and N. Forwood, Judges; H. Jung, Registrar, has given a judgment on 26 October 2004, in which it:',
 'Graphviz': 'digraph {\n    rankdir=LR;\n    node [shape=box];\n\n    "Nicoletta Falcone" -> "M. Condinanzi" [label="represent" dir=none];\n    "The Comission of the European Comminities" -> "Nicoletta Falcone" [label="application for annulment of the decision of 2 May 2002 of the selection board in Competition COM/A/10/01 to exclude the applicant from the written tests on the ground that she did not obtain sufficient marks to be included among the 400 best candidates"];\n}',
 'language': 'English'
}
\end{lstlisting}
}
\end{tabular}
\end{center}

\end{document}

%% file: tables/main_result.tex
\begin{table*}[t]
\centering
\footnotesize
\begin{tabular}{lccccccccc}
\toprule
    &\multicolumn{3}{c}{\textbf{Graph-based}}  & \multicolumn{2}{c}{\textbf{Valid Graph}} & \multicolumn{4}{c}{\textbf{Legal Content}} \\
    \cmidrule(lr){2-4} \cmidrule(lr){5-6} \cmidrule(lr){7-10}
\textbf{Model} &\textbf{G} & \textbf{G-N} & \textbf{G-N-E} & \textbf{Top1} & \textbf{Top10} & \textbf{Entity} & \textbf{R \& T} & \textbf{Source} & \textbf{Statement} \\
\midrule
\multicolumn{7}{l}{\textit{Few-shot results}} \\
CodeLlama 7B & 12.88 & 9.10 & 3.70 & 16.78 & 85.22 & 48.94 & 5.17 & 10.11 & 1.24\\
CodeLlama 7B it. &  15.67 & 11.78 & 6.07 &  37.65 & 89.39 & 55.10 & 8.01 & 11.00 & 1.29 \\
CodeLlama 13B & 15.33 & 10.90 & 5.23 & 17.30 & 85.04 & 51.46 & 7.34 & 11.89 & 2.38\\
CodeLlama 13B it. \hspace{-1em} & 16.37 & 12.35 & 6.47 & 33.39 & 88.70 & 55.00 & 8.54 & 10.76 & 2.21\\
Llama3.1 8B & 26.10 & 20.32 & 11.18 & 30.00 & 83.22 & 64.06 & 14.21 & 16.85 & 2.84\\
Llama3.1 8B it. \hspace{-1em} &  24.47 & 17.91 & 10.32 & 24.00 & 84.00& 62.96 & 13.95 & 16.22 & 2.21\\
Llama3.2 3B & 22.20 & 17.06 & 8.65 & 27.13 & 80.52& 57.35 & 11.18 & 12.24 & 2.28\\
Llama3.2 3B it. \hspace{-1em} &  25.64 & 19.80 & 11.38 & 56.26 & 92.09& 54.11 & 14.51 & 10.93 & 2.78\\
Gemma2 9B  & 15.35 & 11.28 & 5.28 & 35.30 & 93.30& 54.88 & 7.03 & 9.18 & 2.56\\
Gemma2 9B it. & 27.22 & 22.44 & 12.64 & 70.70 & 94.17 & 73.27 & 15.16 & 17.21 & 1.82\\
GPT-3.5-Turbo & 26.66 & 22.28 & 13.51 & 94.26 & \textbf{100.0} & 73.02 & 16.18 & 13.81 & \textbf{3.88}\\
GPT-4         & \textbf{33.46} & \textbf{28.70} & \textbf{19.96} & \textbf{99.13} & \textbf{100.0} & \textbf{75.31} & \textbf{23.24} & \textbf{21.52} & 3.30\\
GPT-4o        & 23.58 & 20.10 & 13.42 & 95.22 & \textbf{100.0} & 75.15 & 15.82 & 19.93 & 2.97\\
\midrule
\multicolumn{7}{l}{\textit{Finetuning results}} \\
CodeLlama 7B &  30.56 & 23.04 & 16.34 & 94.43 & 99.57& 76.73 & 21.54 & 39.81 & 8.59\\
CodeLlama 7B it. \hspace{-1em} & 33.47 & 25.85 & 18.68& 96.61 & 99.65 & 76.90 & 24.00 & 34.61 & 9.03\\
CodeLlama 13B & 34.44 & 25.94 & 17.70 & 97.13 & 99.83& 76.73 & 23.23 & 42.23 & 7.43\\ 
CodeLlama 13B it. \hspace{-1em} & 35.61 & 27.75 & 19.65 & 96.17 & 99.65 & 77.68 & 24.87 & 46.32 & 9.85\\
Llama3.1 8B &  30.09 & 19.86 & 13.25 & 94.70 & \textbf{100.0} & 68.22 & 19.75 & 29.01 & 9.39\\
Llama3.1 8B it. \hspace{-1em} &  29.59 & 20.32 & 13.42 & 87.91 & 99.83& 70.57 & 18.98 & 31.51 & 9.28\\
Llama3.2 3B &  33.38 & 24.29 & 17.56 & 92.78 & 99.83 & 73.29 & 23.83 & 47.47 & 9.89\\
Llama3.2 3B it. \hspace{-1em} &  30.37 & 21.51 & 14.70 & 87.22 & 99.83& 71.93 & 20.38 & 43.24 & 10.08\\
Gemma2 9B  &  \textbf{43.38} & \textbf{36.47} & \textbf{27.52} & \textbf{98.00} & \textbf{100.0}& \textbf{81.85} & \textbf{32.53} & \textbf{50.97} & \textbf{12.75} \\
Gemma2 9B it. &  42.30 & 34.26 & 25.95 & 96.17 & \textbf{100.0}& 81.02 & 31.80 & 42.05 & 11.92\\
\bottomrule
\end{tabular}
\caption{
Overall results of the legal text visualization in the LegalViz test set.
\textbf{G}, \textbf{G-N} and \textbf{G-N-E} denote \texttt{Graph}, \texttt{Graph\&Node} and  \texttt{Graph\&Node\&Edge} respectively. Valid Graph is success rate of creating valid DOT language codes in top-1 and top-10 generated results. ``it.'' means instruct tuning models. 
The highest scores of each column are in bold.
}
\label{table:generation_performance}
\end{table*}

%% file: tables/multi_lang_parts.tex
\begin{table*}[ht]
\setlength{\tabcolsep}{3pt}
  \centering
  \footnotesize
 
  \begin{adjustbox}{max width=\textwidth}
    
    \begin{tabular}{l>{\columncolor{gray!30}}cc>{\columncolor{gray!30}}cc>{\columncolor{gray!30}}cc>{\columncolor{gray!30}}cc>{\columncolor{gray!30}}cc>{\columncolor{gray!30}}cc>{\columncolor{gray!30}}cc>{\columncolor{gray!30}}cc>{\columncolor{gray!30}}cc>{\columncolor{gray!30}}cc>{\columncolor{gray!30}}cc>{\columncolor{gray!30}}cc>{\columncolor{gray!30}}c}
    \toprule
            \textbf{Model} & \textbf{BG} & \textbf{ES} & \textbf{CS} & \textbf{DA} & \textbf{DE} & \textbf{ET} & \textbf{EL} & \textbf{EN} & \textbf{FR} & \textbf{HR} & \textbf{IT} & \textbf{LV} & \textbf{LT} & \textbf{HU} & \textbf{MT} & \textbf{NL} & \textbf{PL} & \textbf{PT} & \textbf{RO} & \textbf{SK} & \textbf{SL} & \textbf{FI} & \textbf{SV} \\
\midrule
\multicolumn{11}{l}{\textit{Entity}} \\
Gemma 2 9B fs. & 59.22 & 59.20 & 52.53 & 54.47 & 56.24 & 53.51 & 53.22 & 59.95 & 53.77 & 55.32 & 55.42 & 51.33 & 55.04 & 42.26 & 47.55 & 60.06 & 51.17 & 61.68 & 59.21 & 52.85 & 50.72 & 55.69 & 57.21\\
Gemma 2 9B ft. & \textbf{80.62} & \textbf{84.33} & \textbf{80.06} & \textbf{82.53} & \textbf{83.31} & \textbf{78.57} & \textbf{80.44} & \textbf{86.98} & \textbf{82.90} & \textbf{81.14} & \textbf{80.94} & \textbf{81.11} & \textbf{77.16} & \textbf{81.66} & \textbf{82.59} & \textbf{82.70} & \textbf{82.58} & \textbf{85.64} & \textbf{83.97} & \textbf{81.27} & \textbf{79.16} & \textbf{79.23} & \textbf{83.22}\\
Avg. fs. models  &  60.78 &  63.32 &  57.12 &  59.67 &  61.30 &  57.67 &  53.59 &  65.77 &  60.61 &  60.48 &  59.06 &  56.85 &  55.53 &  55.73 &  56.64 &  61.37 &  60.18 &  61.36 &  60.89 &  59.50 &  57.20 &  59.68 &  61.04 \\
Avg. ft. models & 73.19  &  77.05  &  73.59  &  74.14  &  74.28  &  72.62  &  72.31  &  78.78  &  75.5  &  74.87  &  75.82  &  71.87  &  72.08  &  72.16  &  73.36  &  75.2  &  74.25  &  76.72  &  76.22  &  74.08  &  72.77  &  72.47  &  75.17\\
\midrule
\multicolumn{11}{l}{\textit{Relations\&Transactions}} \\
Gemma 2 9B fs. & 10.64 &  7.24 &  8.97 & 10.44 &  2.28 &  7.31 &  7.31 & 13.39 &  5.23 &  7.11 &  4.42 &  3.54 &  6.79 &  2.81 &  3.71 &  9.27 &  8.01 & 11.18 &  6.90 &  6.81 &  5.94 &  2.40 &  6.73\\
Gemma 2 9B ft. & \textbf{32.00} & \textbf{33.88} & \textbf{31.19} & \textbf{37.13} & \textbf{32.34} & \textbf{26.13} & \textbf{30.42} & \textbf{36.53} & \textbf{23.06} & \textbf{36.43} & \textbf{27.14} & \textbf{30.74} & \textbf{28.06} & \textbf{38.13} & \textbf{31.88} & \textbf{29.68} & \textbf{40.27} & \textbf{34.39} & \textbf{37.17} & \textbf{35.15} & \textbf{29.50} & \textbf{34.40} & \textbf{33.02}\\
Avg. fs. models  &  11.22 &  11.80 &  10.28 &  13.90 &  11.26 &  8.74 &  12.56 &  10.76 &  11.67 &  10.47 &  10.12 &  12.78 &  12.19 &  11.24 &  9.22 &  13.66 &  12.18 &  11.88 &  10.62 &  10.42 &  11.27 &  10.02 &  11.68 \\
Avg. ft. models & 23.66  &  23.18  &  22.96  &  25.47  &  22.98  &  23.0  &  23.01  &  27.72  &  22.09  &  25.56  &  20.87  &  24.4  &  25.79  &  21.45  &  23.09  &  20.95  &  24.13  &  24.08  &  23.2  &  23.36  &  23.3  &  21.94  &  24.86\\
\midrule
\multicolumn{11}{l}{\textit{Source}} \\
Gemma 2 9B fs. & 0.00 &  0.00 & 15.15 & 10.37 & 13.85 & 17.95 & 15.91 & 17.96 &  0.00 &  9.92 &  0.00 &  4.97 &  0.00 &  6.89 &  0.00 &  6.20 & 19.31 & 12.47 &  6.05 & 14.77 &  0.00 &  7.51 & 23.25\\
Gemma 2 9B ft. & \textbf{58.98} & \textbf{49.07} & \textbf{41.06} & \textbf{52.69} & \textbf{56.19} & \textbf{54.21} & \textbf{47.31} & \textbf{61.32} & \textbf{61.87} & \textbf{51.14} & \textbf{49.90} & \textbf{64.56} & 38.86 & \textbf{42.34} & \textbf{48.10} & \textbf{44.85} & \textbf{53.68} & \textbf{42.08} & \textbf{54.92} & \textbf{48.00} & \textbf{47.03} & \textbf{54.46} & \textbf{54.68}\\
Avg. fs. models  &  12.19 &  11.78 &  14.39 &  13.89 &  13.58 &  12.11 &  12.38 &  14.07 &  15.02 &  15.15 &  13.04 &  14.51 &  14.84 &  13.35 &  11.65 &  16.77 &  14.65 &  13.66 &  12.89 &  15.30 &  9.52 &  11.08 &  15.20 \\
Avg. ft. models & 36.8  &  37.3  &  36.52  &  37.38  &  45.55  &  38.04  &  33.58  &  36.04  &  36.92  &  40.67  &  36.5  &  40.69  &  \textbf{45.54}  &  39.53  &  37.58  &  35.53  &  42.92  &  34.26  &  35.04  &  44.14  &  36.46  &  34.54  &  38.53\\
\midrule
\multicolumn{11}{l}{\textit{Statement}} \\
Gemma 2 9B fs. & 0.00 &  3.25 &  3.27 &  3.74 &  4.42 &  2.30 &  1.53 &  1.54 &  1.59 &  3.85 &  0.00 &  0.00 &  1.95 &  2.41 &  4.40 &  3.94 &  3.03 &  4.86 &  1.40 &  1.32 &  4.43 &  4.54 &  1.45\\
Gemma 2 9B ft. & \textbf{16.78} & \textbf{17.10} &  \textbf{9.17} & \textbf{12.76} &  6.78 &  \textbf{9.86} & \textbf{20.95} &  5.90 & \textbf{11.09} & \textbf{11.13} & \textbf{16.32} & \textbf{11.21} & \textbf{15.90} & \textbf{15.22} & \textbf{11.10} &  \textbf{8.92} & \textbf{12.03} & \textbf{11.90} & \textbf{16.09} & \textbf{17.65} & \textbf{13.17} & \textbf{7.46} & \textbf{14.89} \\
Avg. fs. models  &  2.10 &  2.66 &  1.57 &  2.55 &  2.50 &  1.62 &  2.15 &  3.75 &  2.98 &  1.99 &  1.38 &  2.50 &  3.04 &  1.75 &  1.71 &  2.31 &  3.63 &  2.24 &  2.49 &  1.82 &  2.34 &  1.25 &  1.86 \\
Avg. ft. models & 10.52  &  12.05  &  6.94  &  9.62  &  \textbf{8.64}  &  7.33  &  13.1  &  \textbf{11.23}  &  8.9  &  9.07  &  10.92  &  9.05  &  10.68  &  8.47  &  10.1  &  8.32  &  10.77  &  9.28  &  8.89  &  9.22  &  8.67  &  6.45  &  9.04\\

\bottomrule
\end{tabular}
\end{adjustbox}
\caption{Scores of \texttt{Entity}, \texttt{Relations\&Transactions}, \texttt{Source}, and \texttt{Statement} by 23 languages. ``fs.'' means few-shot and ``ft.'' means finetuned with the LegalViz dataset. Avg. fs. models exclude GPTs for comparison.}
\label{table:scores_by_language}
\end{table*}

%% file: tables/multi_lang_parts_appendix.tex
\begin{table*}[p]
\setlength{\tabcolsep}{3pt}
  \centering
  \footnotesize
  \resizebox{\textwidth}{!}{
    \begin{tabular}{lcccccccccccccccccccccccc}
\toprule
            \textbf{Model} & \textbf{BG} & \textbf{ES} & \textbf{CS} & \textbf{DA} & \textbf{DE} & \textbf{ET} & \textbf{EL} & \textbf{EN} & \textbf{FR} & \textbf{HR} & \textbf{IT} & \textbf{LV} & \textbf{LT} & \textbf{HU} & \textbf{MT} & \textbf{NL} & \textbf{PL} & \textbf{PT} & \textbf{RO} & \textbf{SK} & \textbf{SL} & \textbf{FI} & \textbf{SV} \\
\midrule
\multicolumn{10}{l}{\textit{Few-shot / Test / Entity}} \\
CodeLlama 7B & 50.18 & 55.56 & 42.12 & 52.75 & 58.24 & 45.43 & 41.69 & 60.95 & 50.10 & 55.84 & 47.06 & 45.38 & 42.27 & 45.54 & 45.03 & 44.98 & 46.03 & 53.60 & 45.72 & 39.05 & 40.27 & 50.99 & 52.03\\
CodeLlama 7B it. & 53.99 & 63.38 & 52.62 & 54.18 & 53.87 & 49.99 & 29.70 & 60.82 & 58.77 & 53.09 & 49.93 & 51.03 & 53.94 & 53.87 & 56.61 & 58.57 & 60.00 & 46.15 & 58.42 & 48.33 & 57.92 & 54.25 & 61.59\\
CodeLlama 13B & 49.07 & 55.77 & 46.26 & 53.03 & 46.72 & 55.66 & 36.93 & 62.80 & 52.83 & 54.49 & 50.83 & 44.09 & 46.06 & 49.92 & 43.90 & 49.27 & 50.04 & 56.16 & 46.94 & 51.95 & 56.94 & 56.61 & 49.33\\
CodeLlama 13B it. & 52.34 & 62.72 & 58.54 & 59.43 & 54.41 & 46.90 & 52.79 & 56.59 & 56.90 & 51.12 & 57.74 & 52.98 & 36.93 & 57.23 & 46.21 & 56.67 & 52.92 & 58.54 & 57.48 & 63.31 & 51.15 & 55.77 & 56.61\\
Llama3.1 8B & 66.90 & 68.38 & 58.64 & 57.41 & 65.76 & 64.60 & 63.03 & 66.84 & 67.69 & 69.65 & 58.26 & 59.25 & 53.70 & 58.33 & 57.11 & 69.34 & 62.92 & 67.71 & 64.38 & 69.16 & 61.94 & 68.12 & 67.49\\
Llama3.1 8B it. & 65.82 & 63.81 & 55.71 & 69.35 & 67.13 & 64.15 & 59.84 & 59.17 & 60.19 & 64.38 & 62.66 & 64.62 & 58.35 & 59.65 & 64.43 & 68.54 & 68.94 & 57.49 & 67.28 & 65.32 & 57.56 & 61.16 & 60.17\\
Llama3.2 3B & 60.20 & 59.19 & 57.51 & 58.95 & 58.70 & 54.21 & 55.35 & 66.20 & 51.84 & 56.52 & 55.81 & 53.45 & 58.90 & 52.79 & 58.29 & 59.60 & 60.83 & 58.46 & 57.82 & 56.98 & 50.60 & 54.57 & 59.23\\
Llama3.2 3B it. & 52.65 & 54.02 & 50.29 & 51.98 & 58.38 & 56.44 & 49.93 & 72.07 & 61.23 & 58.34 & 48.55 & 50.38 & 53.42 & 41.90 & 55.48 & 54.39 & 52.91 & 58.18 & 44.34 & 53.30 & 52.59 & 53.11 & 53.78\\
Gemma 2 9B  & 59.22 & 59.20 & 52.53 & 54.47 & 56.24 & 53.51 & 53.22 & 59.95 & 53.77 & 55.32 & 55.42 & 51.33 & 55.04 & 42.26 & 47.55 & 60.06 & 51.17 & 61.68 & 59.21 & 52.85 & 50.72 & 55.69 & 57.21\\
Gemma 2 9B it. & \textbf{76.47} & 75.00 & 71.81 & 71.69 & 74.38 & 64.52 & 70.85 & 77.22 & 72.99 & 73.55 & \textbf{77.12} & 71.24 & 73.37 & 72.69 & 67.58 & 74.73 & 74.18 & 75.99 & \textbf{78.90} & 72.25 & 68.83 & 72.79 & 75.08\\
GPT-3.5-Turbo & 68.54 & 75.68 & \textbf{74.63} & 73.19 & 75.61 & 69.62 & 61.69 & 79.75 & 72.91 & 74.74 & 73.41 & 68.16 & 66.19 & 71.75 & 72.67 & 76.00 & 73.94 & 77.32 & 73.84 & 71.22 & 71.12 & 75.77 & \textbf{79.44}\\
GPT-4 & 74.13 & 75.20 & 73.01 & \textbf{74.59} & 72.73 & \textbf{76.05} & 71.15 & \textbf{82.03} & 75.74 & 73.23 & 76.01 & \textbf{76.25} & 72.66 & 75.75 & \textbf{75.94} & \textbf{77.03} & 73.94 & 76.04 & 76.75 & \textbf{77.52} & \textbf{72.52} & \textbf{76.07} & 78.29\\
GPT-4o & 74.04 & \textbf{77.03} & 72.96 & 69.36 & \textbf{76.60} & 75.63 & \textbf{75.08} & 76.85 & \textbf{77.39} & \textbf{75.48} & 76.83 & 71.09 & \textbf{75.90} & \textbf{79.94} & 73.00 & 76.40 & \textbf{74.50} & \textbf{78.80} & 75.36 & 76.38 & 72.26 & 70.55 & 76.69\\
Avg. fs. (w/ GPTs)  &  63.07 &  65.85 &  60.40 &  62.21 &  64.04 &  60.89 &  56.73 &  68.52 &  63.56 &  63.28 &  62.33 &  59.85 &  58.74 &  59.74 &  60.08 &  64.39 &  62.97 &  64.57 &  63.78 &  62.61 &  60.15 &  62.57 &  64.46 \\
Avg. fs. (w/o GPTs)  &  60.78 &  63.32 &  57.12 &  59.67 &  61.30 &  57.67 &  53.59 &  65.77 &  60.61 &  60.48 &  59.06 &  56.85 &  55.53 &  55.73 &  56.64 &  61.37 &  60.18 &  61.36 &  60.89 &  59.50 &  57.20 &  59.68 &  61.04 \\
\midrule
\multicolumn{10}{l}{\textit{Few-shot / Test / Relations\&Transactions}} \\
CodeLlama 7B & 0.00 &  3.51 &  2.23 &  4.98 &  4.28 &  2.55 &  3.81 &  9.04 &  5.67 &  8.56 &  1.15 &  5.18 &  8.93 &  4.64 &  3.79 &  6.57 &  7.01 &  4.28 &  8.50 &  9.72 &  4.10 &  4.76 &  2.95\\
CodeLlama 7B it. & 8.91 & 12.86 &  9.64 & 13.06 &  8.35 &  3.01 &  4.84 &  4.24 & 11.72 &  2.77 &  6.86 & 11.80 &  4.46 & 10.62 &  4.64 &  9.63 &  8.29 &  5.05 &  5.95 &  4.71 &  7.87 & 13.47 &  8.50\\
CodeLlama 13B & 9.57 &  6.70 &  5.76 &  9.72 &  6.58 &  7.50 &  0.00 &  6.07 &  9.16 &  7.51 &  5.55 &  6.91 &  9.99 &  4.68 &  6.69 &  7.11 &  9.07 &  6.36 &  7.88 &  8.49 & 12.97 &  7.00 &  4.53\\
CodeLlama 13B it. & 6.55 &  7.65 &  8.59 &  9.07 & 16.84 &  7.86 & 12.68 &  4.68 &  4.15 &  5.18 &  5.90 & 12.86 & 10.29 &  8.81 &  6.13 &  8.20 &  8.69 & 13.82 &  7.46 &  2.10 & 15.02 &  8.53 &  9.11\\
Llama3.1 8B & 16.28 & 13.04 & 12.19 & 13.72 & 13.41 &  9.01 & 18.37 &  9.25 & 12.95 & 17.86 &  8.63 & 15.14 & 14.77 & 19.63 & 19.37 & 15.84 & 13.44 & 14.30 & 15.13 & 13.62 & 13.50 & 13.72 & 17.74\\
Llama3.1 8B it. & 15.08 & 12.15 & 14.90 & 17.57 & 15.54 & 11.52 & 14.23 & 11.20 & 10.83 & 11.76 & 17.60 & 15.19 & 19.35 & 15.64 & 11.20 & 22.23 & 11.71 & 16.44 & 11.55 &  8.52 &  9.82 & 10.43 & 16.55\\
Llama3.2 3B & 12.31 & 11.42 &  9.56 & 15.56 &  8.08 &  9.47 & 14.02 & 11.96 & 12.84 & 14.34 &  6.72 & 16.22 & 15.93 & 10.85 &  7.27 & 14.23 & 13.08 &  8.67 &  5.66 & 11.32 &  6.75 &  8.30 & 11.95\\
Llama3.2 3B it. & 9.63 & 13.98 & 14.27 & 16.99 & 11.81 & 13.75 & 15.59 & 14.84 & 13.27 & 15.39 & 12.04 & 13.91 & 15.91 & 14.00 & 10.21 & 20.97 & 17.37 & 14.41 & 13.85 & 14.72 & 16.02 & 12.03 & 16.01\\
Gemma 2 9B  & 10.64 &  7.24 &  8.97 & 10.44 &  2.28 &  7.31 &  7.31 & 13.39 &  5.23 &  7.11 &  4.42 &  3.54 &  6.79 &  2.81 &  3.71 &  9.27 &  8.01 & 11.18 &  6.90 &  6.81 &  5.94 &  2.40 &  6.73\\
Gemma 2 9B it. & 16.45 & 17.92 & 14.25 & 17.10 & 11.44 & 10.98 & \textbf{23.63} & 13.33 & 20.05 & 10.55 & 17.30 & 17.09 & 12.40 & 13.66 & 13.30 & 16.58 & 17.31 & 12.70 & 15.26 & 14.44 & 18.21 & 13.19 & 11.66\\
GPT-3.5-Turbo & 15.28 & 19.60 & 13.99 & 19.73 & 18.93 & 15.10 & 11.90 & 14.27 & 14.99 & 13.10 & 16.13 & 13.24 & \textbf{18.35} & 17.74 & 16.32 & 20.83 & 12.90 & 17.05 & 22.55 & 15.89 & 13.53 & 14.13 & 18.83\\
GPT-4 & 17.57 & \textbf{24.20} & \textbf{21.37} & \textbf{22.72} & \textbf{23.62} & \textbf{20.48} & 21.71 & \textbf{21.71} & \textbf{24.75} & \textbf{21.89} & \textbf{22.88} & \textbf{27.54} & 16.42 & \textbf{25.90} & \textbf{20.87} & \textbf{23.79} & \textbf{25.73} & \textbf{25.02} & \textbf{23.31} & \textbf{27.87} & \textbf{25.55} & \textbf{20.12} & \textbf{32.75}\\
GPT-4o & \textbf{18.03} & 19.58 & 17.97 & 11.52 & 15.15 & 17.90 & 17.59 & 18.19 & 15.00 & 16.24 & 14.32 & 14.82 & 16.92 & 20.24 & 10.93 & 17.20 & 15.76 & 13.83 & 19.80 & 11.12 & 15.57 & 15.17 & 12.70\\
Avg. fs. (w/ GPTs)  &  12.37 &  13.66 &  11.78 &  14.72 &  12.86 &  10.56 &  13.46 &  12.22 &  12.98 &  11.79 &  11.66 &  13.93 &  13.20 &  13.25 &  10.59 &  15.05 &  13.37 &  13.23 &  12.87 &  12.00 &  12.66 &  11.31 &  13.63 \\
Avg. fs. (w/o GPTs)  &  11.22 &  11.80 &  10.28 &  13.90 &  11.26 &  8.74 &  12.56 &  10.76 &  11.67 &  10.47 &  10.12 &  12.78 &  12.19 &  11.24 &  9.22 &  13.66 &  12.18 &  11.88 &  10.62 &  10.42 &  11.27 &  10.02 &  11.68 \\
\midrule
\multicolumn{10}{l}{\textit{Few-shot / Test / Source}} \\
CodeLlama 7B & 6.17 &  5.95 &  9.52 & 11.17 & 11.81 &  5.88 &  0.00 &  9.76 &  7.98 & 13.35 & 14.04 &  0.00 &  8.78 & 14.63 &  0.00 &  9.36 & 17.10 &  7.97 &  5.30 & 19.42 &  0.00 & 11.84 & \textbf{23.50}\\
CodeLlama 7B it. & 12.59 &  9.49 &  6.55 &  3.73 &  8.81 &  8.48 &  0.00 & 23.01 &  3.23 &  4.24 &  6.09 & 17.16 & 19.91 & 15.25 & 18.50 & 32.75 &  3.62 & 14.67 &  8.11 & 14.48 &  8.39 &  6.66 &  7.35\\
CodeLlama 13B & 8.52 & 15.86 &  0.00 & 12.15 & 16.66 &  9.48 & \textbf{62.75} &  3.30 & 16.30 & 19.20 & 11.36 & 19.90 & 19.90 &  3.23 & 13.68 &  8.81 &  5.16 & 10.66 & 12.65 & 18.01 & 13.07 &  9.99 &  6.71\\
CodeLlama 13B it. & 19.31 &  8.48 & \textbf{26.78} & 11.31 &  3.83 &  4.67 &  0.00 & 11.44 &  9.84 & 19.79 &  9.16 & 13.01 &  9.16 &  4.88 &  8.44 & 11.55 &  5.26 &  9.40 & 12.98 &  9.28 &  4.85 & 12.13 & 14.57\\
Llama3.1 8B & 21.36 & \textbf{23.52} & 16.62 & 12.39 & 18.54 &  6.89 & 14.75 & 22.88 & \textbf{30.57} & 11.37 & 10.62 & \textbf{29.81} &  6.44 & 16.97 & 12.50 & 17.37 & 24.46 & 14.77 & 19.83 & 17.49 &  2.81 & 19.26 & 18.86\\
Llama3.1 8B it. & 19.37 & 11.92 &  8.42 & 19.98 & 20.55 & 18.02 &  7.13 & 13.90 & 25.90 & 22.06 & \textbf{20.09} & 18.24 & \textbf{30.41} & 15.70 & 12.04 & 26.56 & 11.16 & 11.98 & 10.87 & 10.98 & 12.90 &  8.47 & 18.45\\
Llama3.2 3B & 2.74 &  4.90 & 21.67 & 13.32 & 16.26 &  5.88 & 11.79 &  9.30 & 14.26 & 15.43 & 10.81 &  4.08 & 12.55 & 23.42 & 10.77 & 24.08 &  7.41 & 12.88 & 12.66 & 16.80 & 10.84 &  8.79 &  4.55\\
Llama3.2 3B it. & 6.04 & 10.96 & 15.08 & 12.65 & 15.50 & 10.11 & 13.34 &  6.11 & 17.62 &  8.69 & 10.48 &  9.27 & 15.68 & 11.42 &  5.14 &  4.44 &  9.77 & 14.28 & 12.53 &  5.51 & 10.13 & 11.10 & 10.54\\
Gemma 2 9B  & 0.00 &  0.00 & 15.15 & 10.37 & 13.85 & 17.95 & 15.91 & 17.96 &  0.00 &  9.92 &  0.00 &  4.97 &  0.00 &  6.89 &  0.00 &  6.20 & 19.31 & 12.47 &  6.05 & 14.77 &  0.00 &  7.51 & 23.25\\
Gemma 2 9B it. & 19.39 &  9.11 & 15.75 & 19.30 & 17.24 & 23.81 &  6.92 & 19.14 & 19.80 & 17.52 & 18.29 & 17.30 & 13.52 & 17.44 &  5.64 & 19.66 & \textbf{28.24} & 10.95 & 18.68 & 19.81 & 21.13 & 12.12 & 17.57\\
GPT-3.5-Turbo & 16.08 &  9.12 &  6.07 & 16.56 &  8.26 & 19.92 &  8.41 & 14.33 & 13.69 &  8.54 & 13.55 & 11.80 & 15.82 & 10.99 & \textbf{28.10} & 18.19 & 16.02 &  9.26 & 18.24 &  6.61 & 12.65 & \textbf{25.69} & 10.43\\
GPT-4 & 16.34 & 13.81 & 20.70 & \textbf{22.26} & \textbf{24.45} & \textbf{27.83} & 33.69 & \textbf{25.31} & 25.86 & 22.81 & 17.29 & 19.08 & 23.43 & 15.76 & 18.60 & \textbf{37.61} & 11.85 & \textbf{24.25} & 17.20 & 14.73 & \textbf{24.94} & 18.70 & 19.64\\
GPT-4o & \textbf{22.13} & 17.89 & 16.58 & 20.44 & 17.37 & 19.33 & 25.70 & 23.88 & 19.32 & \textbf{23.99} & 17.77 & 22.92 & 24.56 & \textbf{24.05} & 13.98 & 18.48 & 17.70 & 20.11 & \textbf{23.32} & 17.95 & 15.01 & 18.93 & 21.76\\
Avg. fs. (w/ GPTs)  &  13.39 &  12.14 &  14.40 &  15.06 &  14.20 &  14.16 &  14.43 &  15.49 &  15.94 &  15.81 &  13.67 &  15.20 &  16.12 &  14.07 &  13.37 &  18.37 &  14.76 &  14.50 &  14.23 &  14.86 &  11.13 &  13.09 &  15.62 \\
Avg. fs. (w/o GPTs)  &  12.19 &  11.78 &  14.39 &  13.89 &  13.58 &  12.11 &  12.38 &  14.07 &  15.02 &  15.15 &  13.04 &  14.51 &  14.84 &  13.35 &  11.65 &  16.77 &  14.65 &  13.66 &  12.89 &  15.30 &  9.52 &  11.08 &  15.20 \\
\midrule
\multicolumn{10}{l}{\textit{Few-shot / Test / Statement}} \\
CodeLlama 7B & 2.30 &  4.56 &  0.00 &  0.00 &  1.68 &  0.00 &  \textbf{6.76} &  0.00 &  0.00 &  3.17 &  0.00 &  0.00 &  4.10 &  0.00 &  1.58 &  0.00 &  3.10 &  2.14 &  0.00 &  0.00 &  0.00 &  0.00 &  1.79 \\
CodeLlama 7B it. & 3.71 &  4.02 &  0.00 &  0.00 &  2.02 &  0.00 &  0.00 &  1.52 &  4.14 &  2.38 &  0.00 &  1.95 &  0.00 &  0.00 &  0.00 &  0.00 &  5.42 &  0.00 &  0.00 &  0.00 &  0.00 &  0.00 &  1.93\\
CodeLlama 13B & 1.77 &  0.00 &  2.16 &  4.05 &  4.55 &  2.68 &  0.00 &  3.36 &  3.16 &  0.00 &  \textbf{4.18} &  1.82 &  2.18 &  0.00 &  4.28 &  3.79 &  3.67 &  1.88 &  1.83 &  0.00 &  3.26 &  2.26 &  0.00\\
CodeLlama 13B it. & 0.00 &  4.12 &  1.67 &  7.71 &  2.99 &  2.10 &  0.00 &  3.93 &  \textbf{4.36} &  1.74 &  0.00 &  1.61 &  2.05 &  1.82 &  0.00 &  2.00 &  2.43 &  1.22 &  2.26 &  1.64 &  3.35 &  0.00 &  0.00\\
Llama3.1 8B & 4.30 &  1.61 &  1.33 &  0.00 &  1.40 &  0.00 &  3.70 &  7.91 &  3.60 &  1.89 &  0.00 &  2.45 &  6.76 &  \textbf{4.83} &  0.00 &  1.84 &  4.89 &  3.00 &  4.84 &  1.78 &  4.41 &  1.54 &  3.09\\
Llama3.1 8B it. & 1.88 &  2.36 &  0.00 &  \textbf{7.72} & \textbf{6.25} &  1.46 &  0.00 &  3.47 &  2.35 &  0.00 &  0.00 &  \textbf{7.51} &  2.67 &  0.00 &  0.00 &  2.93 &  2.48 &  0.00 &  1.23 &  4.71 &  4.50 &  0.00 &  0.00\\
Llama3.2 3B & 0.00 &  0.00 &  1.54 &  0.00 &  0.00 &  3.94 &  2.73 &  5.66 &  3.84 &  1.79 &  1.47 &  0.00 &  3.91 &  3.42 &  2.36 &  3.63 &  3.58 &  3.49 &  4.62 &  2.47 &  0.00 &  0.00 &  3.45\\
Llama3.2 3B it. & \textbf{5.05} &  2.15 &  \textbf{4.07} &  2.94 &  0.00 &  \textbf{4.16} &  0.00 &  7.31 &  3.83 &  3.80 &  0.00 &  2.50 &  0.00 &  0.00 &  2.84 &  0.00 &  \textbf{7.57} &  3.61 &  3.08 &  3.71 &  2.04 &  0.00 &  3.55\\
Gemma 2 9B  & 0.00 &  3.25 &  3.27 &  3.74 &  4.42 &  2.30 &  1.53 &  1.54 &  1.59 &  3.85 &  0.00 &  0.00 &  1.95 &  2.41 &  \textbf{4.40} &  3.94 &  3.03 &  \textbf{4.86} &  1.40 &  1.32 &  4.43 &  4.54 &  1.45\\
Gemma 2 9B it. & 0.00 &  1.65 &  2.16 &  0.00 &  0.00 &  0.00 &  1.92 &  1.62 &  1.93 &  1.95 &  1.84 &  2.59 &  \textbf{6.87} &  0.00 &  0.00 &  \textbf{4.71} &  2.30 &  0.00 &  4.12 &  3.03 &  0.00 &  2.34 &  2.08\\
GPT-3.5-Turbo & 2.11 & \textbf{14.42} &  3.19 &  4.67 &  2.78 &  1.92 &  5.26 &  \textbf{9.75} &  4.11 &  0.00 &  0.00 &  2.89 &  3.78 &  2.17 &  4.36 &  1.40 &  1.16 &  2.33 &  1.69 &  \textbf{7.58} &  2.38 &  \textbf{6.04} &  4.46\\
GPT-4 & 1.53 &  0.00 &  2.27 &  1.68 &  4.46 &  0.00 &  3.77 &  4.34 &  2.25 &  \textbf{5.12} &  1.89 &  2.38 &  1.82 &  1.89 &  2.19 &  1.80 &  1.02 &  0.00 &  4.73 &  1.87 &  \textbf{7.83} &  4.08 & \textbf{15.05}\\
GPT-4o & 3.41 &  5.84 &  1.52 &  0.00 &  1.25 &  0.00 &  3.62 &  6.39 &  1.65 &  4.41 &  2.50 &  1.09 &  4.85 &  2.18 &  2.71 &  0.00 &  4.16 &  4.57 &  \textbf{6.36} &  4.89 &  1.63 &  3.24 &  3.18\\
Avg. fs. (w/ GPTs)  &  2.15 &  3.48 &  1.72 &  2.46 &  2.57 &  1.42 &  2.56 &  4.36 &  2.92 &  2.23 &  1.40 &  2.42 &  3.13 &  1.81 &  1.98 &  2.06 &  3.33 &  2.25 &  2.84 &  2.41 &  2.66 &  1.89 &  3.00 \\
Avg. fs. (w/o GPTs)  &  2.10 &  2.66 &  1.57 &  2.55 &  2.50 &  1.62 &  2.15 &  3.75 &  2.98 &  1.99 &  1.38 &  2.50 &  3.04 &  1.75 &  1.71 &  2.31 &  3.63 &  2.24 &  2.49 &  1.82 &  2.34 &  1.25 &  1.86 \\
\bottomrule
\end{tabular}
    }
    \caption{Scores of \texttt{Entity}, \texttt{Relations\&Transactions}, \texttt{Source}, and \texttt{Statement} by 23 languages in EUR-LEX with \textit{few-shot} models.
    }
    \label{table:scores_by_language_all_fewshot}
\end{table*}

\begin{table*}[p]
\setlength{\tabcolsep}{3pt}
  \centering
  \footnotesize
  \resizebox{\textwidth}{!}{
    \begin{tabular}{lcccccccccccccccccccccccc}
\toprule
            \textbf{Model} & \textbf{BG} & \textbf{ES} & \textbf{CS} & \textbf{DA} & \textbf{DE} & \textbf{ET} & \textbf{EL} & \textbf{EN} & \textbf{FR} & \textbf{HR} & \textbf{IT} & \textbf{LV} & \textbf{LT} & \textbf{HU} & \textbf{MT} & \textbf{NL} & \textbf{PL} & \textbf{PT} & \textbf{RO} & \textbf{SK} & \textbf{SL} & \textbf{FI} & \textbf{SV} \\
\midrule

\multicolumn{11}{l}{\textit{Finetuning / Test / Entity}} \\
CodeLlama 7B & 73.32 & 81.37 & 77.15 & 75.90 & 77.03 & 72.49 & 70.26 & 81.64 & 78.65 & 79.09 & 77.07 & 74.27 & 74.65 & 76.25 & 76.06 & 79.70 & 78.15 & 79.80 & 80.26 & 72.35 & 73.67 & 73.12 & 81.34\\
CodeLlama 7B it. & 75.44 & 79.71 & 72.32 & 78.07 & 77.51 & 72.54 & 70.55 & 81.57 & 82.62 & 78.77 & 79.72 & 74.49 & 76.01 & 75.89 & 75.19 & 76.98 & 74.88 & 80.81 & 79.89 & 76.53 & 73.88 & 75.53 & 78.87\\
CodeLlama 13B & 73.57 & 80.71 & 76.43 & 77.41 & 76.64 & 76.17 & 71.35 & 84.57 & 81.59 & 75.94 & 78.37 & 71.78 & 70.45 & 77.05 & 72.85 & 78.20 & 77.02 & 80.27 & 80.37 & 75.75 & 75.25 & 73.97 & 77.78\\
CodeLlama 13B it. & 80.11 & 78.71 & 78.83 & 80.13 & 76.19 & 73.77 & 68.05 & 83.94 & 77.68 & 77.97 & 75.45 & 77.05 & 71.45 & 75.11 & 77.41 & 81.75 & 80.58 & 79.14 & 80.69 & 78.34 & 79.49 & 76.19 & 77.66\\
Llama3.1 8B & 62.97 & 72.32 & 67.14 & 65.75 & 66.59 & 68.89 & 73.31 & 72.38 & 68.46 & 69.80 & 74.28 & 67.21 & 68.32 & 59.81 & 67.13 & 68.65 & 70.81 & 68.77 & 68.79 & 68.14 & 68.89 & 64.29 & 66.85\\
Llama3.1 8B it. & 70.36 & 69.29 & 70.56 & 72.34 & 73.01 & 71.06 & 66.90 & 74.77 & 71.87 & 69.86 & 70.13 & 69.74 & 69.42 & 69.83 & 70.16 & 70.03 & 70.89 & 70.97 & 71.38 & 70.57 & 67.24 & 70.28 & 72.42\\
Llama3.2 3B & 72.37 & 77.53 & 73.46 & 73.95 & 67.78 & 69.34 & 72.43 & 81.45 & 72.44 & 72.39 & 74.98 & 69.92 & 75.28 & 69.13 & 71.96 & 74.68 & 75.52 & 74.32 & 75.71 & 74.96 & 72.20 & 70.22 & 74.31\\
Llama3.2 3B it. & 70.63 & 73.24 & 70.98 & 70.51 & 72.90 & 74.40 & 74.27 & 77.95 & 72.38 & 74.68 & 73.21 & 65.35 & 70.67 & 70.14 & 72.06 & 74.77 & 68.43 & 72.92 & 69.22 & 72.61 & 67.06 & 70.91 & 75.39\\
Gemma 2 9B  & \textbf{80.62} & \textbf{84.33} & 80.06 & \textbf{82.53} & \textbf{83.31} & \textbf{78.57} & \textbf{80.44} & \textbf{86.98} & \textbf{82.90} & \textbf{81.14} & 80.94 & \textbf{81.11} & 77.16 & \textbf{81.66} & \textbf{82.59} & \textbf{82.70} & \textbf{82.58} & \textbf{85.64} & \textbf{83.97} & 81.27 & 79.16 & 79.23 & \textbf{83.22}\\
Gemma 2 9B it. & 78.99 & 83.68 & \textbf{80.78} & 81.75 & 81.71 & 78.33 & 78.59 & 86.25 & 82.09 & 79.06 & \textbf{83.72} & 79.63 & \textbf{79.36} & 80.43 & 76.61 & 82.49 & 81.26 & 82.41 & 82.53 & \textbf{82.70} & \textbf{81.53} & \textbf{79.81} & 79.58\\
Avg. of all ft. & 73.19  &  77.05  &  73.59  &  74.14  &  74.28  &  72.62  &  72.31  &  78.78  &  75.5  &  74.87  &  75.82  &  71.87  &  72.08  &  72.16  &  73.36  &  75.2  &  74.25  &  76.72  &  76.22  &  74.08  &  72.77  &  72.47  &  75.17\\
\midrule
\multicolumn{11}{l}{\textit{Finetuning / Test / Relations\&Transactions}} \\
CodeLlama 7B & 20.37 & 23.62 & 20.02 & 23.86 & 20.28 & 22.99 & 17.96 & 27.69 & 19.03 & 28.93 & 18.38 & 21.97 & 24.81 & 20.11 & 24.10 & 17.07 & 23.34 & 26.13 & 17.04 & 18.47 & 17.15 & 18.38 & 21.80\\
CodeLlama 7B it. & 24.76 & 25.94 & 21.52 & 22.09 & 26.63 & 23.13 & 16.73 & 30.10 & 24.87 & 26.81 & 21.02 & 27.64 & 25.65 & 25.66 & 25.49 & 18.24 & 25.56 & 23.10 & 20.69 & 25.55 & 23.26 & 18.78 & 27.49\\
CodeLlama 13B & 24.21 & 26.41 & 19.90 & 26.90 & 25.74 & 20.58 & 22.42 & 29.09 & 25.10 & 23.05 & 22.43 & 23.17 & 25.01 & 19.15 & 22.66 & 18.23 & 23.07 & 21.53 & 29.45 & 17.11 & 24.34 & 22.06 & 22.70\\
CodeLlama 13B it. & 25.54 & 27.40 & 27.98 & 25.54 & 24.11 & 17.39 & 17.23 & 29.34 & 23.37 & 27.79 & 20.93 & 24.73 & 27.89 & 21.61 & 23.51 & 22.86 & 30.31 & 21.96 & 23.45 & 30.25 & 23.45 & 22.73 & 29.24\\
Llama3.1 8B & 19.39 & 21.58 & 16.94 & 23.98 & 19.14 & 18.17 & 18.67 & 23.87 & 17.46 & 24.37 & 25.34 & 22.22 & 22.60 & 15.41 & 24.38 & 15.50 & 19.78 & 20.40 & 12.84 & 21.98 & 19.49 & 14.94 & 14.48\\
Llama3.1 8B it. & 24.59 & 15.53 & 20.03 & 18.09 & 18.98 & 22.66 & 25.37 & 25.03 & 18.24 & 17.76 & 11.67 & 20.96 & 19.35 & 13.05 & 15.60 & 12.49 & 20.64 & 19.88 & 17.39 & 19.87 & 17.68 & 19.88 & 21.04\\
Llama3.2 3B & 23.24 & 23.33 & 26.23 & 26.27 & 20.87 & 26.10 & 31.03 & 27.78 & 20.06 & 20.31 & 18.46 & 25.99 & 26.80 & 22.12 & 22.60 & 25.05 & 22.67 & 21.78 & 26.80 & 21.74 & 19.74 & 24.09 & 24.45\\
Llama3.2 3B it. & 16.87 & 16.28 & 19.79 & 19.97 & 18.43 & 20.87 & 14.05 & 25.24 & 19.75 & 27.29 & 15.77 & 22.22 & 25.57 & 17.40 & 20.68 & 19.17 & 16.75 & 20.98 & 21.30 & 23.20 & 22.57 & 20.08 & 24.40\\
Gemma 2 9B  & 32.00 & \textbf{33.88} & 31.19 & \textbf{37.13} & \textbf{32.34} & 26.13 & 30.42 & \textbf{36.53} & 23.06 & \textbf{36.43} & 27.14 & 30.74 & 28.06 & \textbf{38.13} & \textbf{31.88} & 29.68 & \textbf{40.27} & \textbf{34.39} & \textbf{37.17} & \textbf{35.15} & 29.50 & \textbf{34.40} & 33.02\\
Gemma 2 9B it. & \textbf{33.17} & 29.21 & \textbf{31.27} & 35.75 & 28.81 & \textbf{31.69} & \textbf{32.34} & 36.41 & \textbf{34.80} & 31.75 & \textbf{29.01} & \textbf{32.96} & \textbf{32.28} & 28.43 & 23.78 & \textbf{31.48} & 32.91 & 33.25 & 32.43 & 26.63 & \textbf{32.69} & 32.06 & \textbf{38.59}\\
Avg. of all ft. & 23.66  &  23.18  &  22.96  &  25.47  &  22.98  &  23.0  &  23.01  &  27.72  &  22.09  &  25.56  &  20.87  &  24.4  &  25.79  &  21.45  &  23.09  &  20.95  &  24.13  &  24.08  &  23.2  &  23.36  &  23.3  &  21.94  &  24.86\\
\midrule
\multicolumn{11}{l}{\textit{Finetuning / Test / Source}} \\
CodeLlama 7B & 34.38 & 43.62 & 35.77 & 46.13 & 44.82 & 25.75 & 21.75 & 39.97 & 29.37 & 40.54 & 40.29 & 47.05 & 46.47 & 37.41 & 37.97 & 34.13 & \textbf{62.24} & 40.33 & 39.16 & 51.49 & 38.13 & 31.97 & 40.31\\
CodeLlama 7B it. & 32.41 & 32.82 & 31.85 & 37.11 & 39.94 & 25.15 & 46.10 & 44.39 & 21.17 & 52.46 & 20.08 & 35.18 & 42.35 & 45.04 & \textbf{54.77} & 22.67 & 31.66 & 43.99 & 16.26 & 36.93 & 19.48 & 29.95 & 28.17\\
CodeLlama 13B & 35.65 & 43.53 & 41.15 & 46.18 & 63.52 & 43.24 & 34.41 & 24.27 & 44.83 & 33.90 & 28.60 & 54.66 & 52.82 & 50.99 & 38.40 & 42.08 & 34.04 & 32.67 & 49.80 & 41.97 & 47.46 & 30.17 & 48.60\\
CodeLlama 13B it. & 40.02 & 43.29 & 47.03 & 43.57 & 43.33 & 48.47 & 40.50 & 50.79 & 57.18 & 44.94 & 37.68 & 52.05 & 47.81 & 40.67 & 46.14 & 44.28 & 58.42 & \textbf{60.51} & 33.88 & 51.98 & 46.54 & 39.50 & 50.57\\
Llama3.1 8B & 23.01 & 22.16 & 22.15 & 20.86 & 28.16 & 24.34 & 41.27 & 47.63 & 37.55 & 34.85 & 41.64 & 33.21 & 38.47 & 21.17 & 19.76 & 33.12 & 34.45 & 21.32 & 25.88 & 34.69 & 24.58 & 15.03 & 18.73\\
Llama3.1 8B it. & 32.66 & 14.99 & 20.08 & 13.07 & \textbf{63.17} & 31.72 & 23.86 & 39.88 & 26.47 & 32.90 & 38.27 & 39.39 & 31.04 & 28.50 & 32.16 & 23.62 & 39.99 & 28.37 & 33.40 & 53.18 & 28.52 & 25.32 & 26.68\\
Llama3.2 3B & 44.74 & 42.78 & 44.67 & 33.53 & 50.74 & 48.70 & 36.06 & 28.56 & 47.22 & \textbf{55.75} & 46.30 & 42.73 & 52.85 & 47.73 & 53.06 & \textbf{53.97} & 54.84 & 36.77 & 48.11 & \textbf{63.84} & \textbf{50.23} & 44.27 & \textbf{56.59}\\
Llama3.2 3B it. & 46.08 & 45.33 & 44.05 & 44.85 & 47.58 & 47.82 & \textbf{48.14} & 24.18 & 36.22 & 46.93 & 36.98 & 37.18 & \textbf{59.37} & \textbf{55.39} & 40.93 & 29.30 & 41.39 & 43.34 & 33.41 & 50.93 & 35.96 & 41.85 & 49.15\\
Gemma 2 9B  & \textbf{58.98} & 49.07 & 41.06 & \textbf{52.69} & 56.19 & \textbf{54.21} & 47.31 & \textbf{61.32} & \textbf{61.87} & 51.14 & \textbf{49.90} & \textbf{64.56} & 38.86 & 42.34 & 48.10 & 44.85 & 53.68 & 42.08 & \textbf{54.92} & 48.00 & 47.03 & \textbf{54.46} & 54.68\\
Gemma 2 9B it. & 36.66 & \textbf{49.94} & \textbf{51.80} & 50.35 & 40.58 & 38.19 & 35.05 & 36.77 & 34.30 & 34.66 & 34.26 & 48.73 & 48.78 & 45.72 & 41.29 & 49.44 & 50.11 & 37.17 & 51.96 & 37.41 & 36.27 & 42.39 & 39.58\\
Avg. of all ft. & 36.8  &  37.3  &  36.52  &  37.38  &  45.55  &  38.04  &  33.58  &  36.04  &  36.92  &  40.67  &  36.5  &  40.69  &  45.54  &  39.53  &  37.58  &  35.53  &  42.92  &  34.26  &  35.04  &  44.14  &  36.46  &  34.54  &  38.53\\
\midrule
\multicolumn{11}{l}{\textit{Finetuning / Test / Statement}} \\
CodeLlama 7B & 6.71 & 10.95 &  5.57 & 13.32 &  6.49 &  6.39 &  9.22 & 11.44 &  6.80 &  9.58 & 10.20 &  9.97 & 11.04 &  8.18 &  7.11 & \textbf{13.23} &  9.88 &  8.58 &  8.36 &  6.04 &  6.59 &  5.37 &  6.58\\
CodeLlama 7B it. & 8.60 & 12.80 &  4.81 &  5.31 &  7.39 &  6.40 &  8.93 & 16.28 &  6.89 & 12.20 &  7.11 & 10.06 & 10.09 &  9.27 &  8.86 &  7.62 & 11.28 & 10.17 &  8.28 & 12.75 &  6.24 &  \textbf{8.95} &  6.57\\
CodeLlama 13B & 4.56 & 11.85 &  4.25 &  7.57 &  9.41 &  7.27 & 11.12 &  5.38 & 11.55 &  8.99 & 15.85 &  4.81 &  8.30 &  8.06 &  8.92 &  8.21 &  9.44 &  4.54 &  4.17 &  0.00 &  5.27 &  3.97 &  7.63\\
CodeLlama 13B it. & 13.87 & 12.88 &  5.71 & 14.42 &  7.89 & \textbf{10.89} &  5.98 & 10.41 &  8.82 & 11.20 &  6.35 & 13.09 & 10.02 &  5.55 &  9.60 & 10.24 & 12.18 & 11.04 &  9.05 & 11.85 &  9.13 &  8.54 &  6.83\\
Llama3.1 8B & 4.77 & 15.59 &  6.58 & \textbf{14.82} &  6.95 &  1.99 & 17.43 &  8.75 &  8.99 &  6.58 &  9.60 &  9.59 &  8.26 &  8.92 & 13.35 &  9.82 & \textbf{17.03} &  8.44 & 10.64 &  6.16 &  7.00 &  4.57 &  9.65\\
Llama3.1 8B it. & 13.77 &  5.91 &  7.47 &  4.46 &  9.11 &  6.62 & 15.23 & \textbf{16.85} &  7.67 &  8.04 &  9.90 &  9.69 &  7.47 &  6.65 &  7.61 &  3.66 & 12.22 &  9.76 & 11.23 & 10.62 & 10.46 &  8.91 &  7.60\\
Llama3.2 3B & 13.19 & 13.40 &  8.12 &  7.09 &  8.80 &  8.32 & 18.02 & 14.03 &  9.31 &  7.49 & 11.53 &  4.74 & 12.58 &  5.44 & \textbf{15.22} &  9.71 &  9.73 &  5.10 &  7.69 &  9.54 & 11.65 &  3.49 & 12.02\\
Llama3.2 3B it. & 10.48 & \textbf{17.49} &  7.43 &  9.56 &  8.96 & 10.34 & 17.02 & 15.98 &  5.35 & \textbf{12.34} & 12.50 &  4.87 & 12.31 &  8.67 & 14.52 &  6.18 &  6.97 &  9.81 &  6.18 &  7.13 &  7.62 &  8.82 & 10.35\\
Gemma 2 9B  & \textbf{16.78} & 17.10 &  9.17 & 12.76 &  6.78 &  9.86 & \textbf{20.95} &  5.90 & 11.09 & 11.13 & \textbf{16.32} & 11.21 & 15.90 & \textbf{15.22} & 11.10 &  8.92 & 12.03 & 11.90 & \textbf{16.09} & \textbf{17.65} & \textbf{13.17} &  7.46 & 14.89\\
Gemma 2 9B it. & 11.97 &  9.83 & \textbf{11.92} & 11.53 & \textbf{14.05} &  8.11 &  9.28 & 13.28 & \textbf{14.06} &  8.85 &  8.91 & \textbf{15.89} & \textbf{16.12} &  9.60 &  8.93 &  8.04 & 14.34 & \textbf{17.06} & 13.05 & 14.82 & 10.83 &  7.00 & \textbf{16.67}\\
Avg. of all ft. & 10.52  &  12.05  &  6.94  &  9.62  &  8.64  &  7.33  &  13.1  &  11.23  &  8.9  &  9.07  &  10.92  &  9.05  &  10.68  &  8.47  &  10.1  &  8.32  &  10.77  &  9.28  &  8.89  &  9.22  &  8.67  &  6.45  &  9.04\\
\bottomrule
\end{tabular}
    }
    \caption{Scores of \texttt{Entity}, \texttt{Relations\&Transactions}, \texttt{Source}, and \texttt{Statement} by 23 languages in EUR-LEX  with \textit{fine-tuned} models.
    }
    \label{table:scores_by_language_all_finetuned}
\end{table*}

%% file: tables/multi_lang_gne_appendix.tex
\begin{table*}[p]
\setlength{\tabcolsep}{3pt}
  \centering
  \footnotesize
  \resizebox{\textwidth}{!}{
    \begin{tabular}{lcccccccccccccccccccccccc}
\toprule
            \textbf{Model} & \textbf{BG} & \textbf{ES} & \textbf{CS} & \textbf{DA} & \textbf{DE} & \textbf{ET} & \textbf{EL} & \textbf{EN} & \textbf{FR} & \textbf{HR} & \textbf{IT} & \textbf{LV} & \textbf{LT} & \textbf{HU} & \textbf{MT} & \textbf{NL} & \textbf{PL} & \textbf{PT} & \textbf{RO} & \textbf{SK} & \textbf{SL} & \textbf{FI} & \textbf{SV} \\
\midrule
\multicolumn{10}{l}{\textit{Few-shot / Test / Graph}} \\
CodeLlama 7B & 6.74 & 19.20 &  5.83 & 10.43 & 12.07 &  8.55 & 11.76 & 20.00 & 13.79 & 15.91 &  5.31 & 16.07 & 18.18 &  9.20 &  6.67 & 13.08 & 15.38 & 12.60 & 17.48 & 19.30 & 12.20 &  6.82 & 12.24\\
CodeLlama 7B it. & 13.56 & 26.36 & 17.31 & 22.58 & 12.50 & 12.50 &  7.69 & 12.00 & 23.78 & 12.63 &  8.16 & 16.22 &  9.52 & 21.78 &  6.38 & 20.87 & 14.43 & 16.30 & 13.91 & 16.07 & 11.21 & 22.50 & 15.00\\
CodeLlama 13B & 18.02 & 12.21 & 11.67 & 19.51 & 12.86 & 17.48 &  0.00 & 16.99 & 16.90 & 12.28 & 16.39 & 19.51 & 19.13 &  7.09 & 11.76 & 18.80 & 18.97 & 17.91 & 15.50 & 19.23 & 19.86 & 12.50 & 13.16\\
CodeLlama 13B it. & 8.51 & 17.78 & 13.79 & 16.79 & 25.95 & 15.52 & 23.08 & 10.39 &  7.63 & 17.65 & 13.04 & 26.79 & 22.03 & 15.38 & 11.97 & 14.06 & 16.67 & 27.07 & 13.73 &  8.63 & 25.71 & 11.65 & 16.15\\
Llama3.1 8B & 26.35 & 26.46 & 21.47 & 22.89 & 22.95 & 22.62 & 35.14 & 19.79 & 25.93 & \textbf{32.56} & 16.48 & 21.21 & 26.23 & 32.84 & 29.41 & 29.14 & 22.36 & 29.35 & 26.09 & 25.30 & 27.38 & \textbf{31.34} & 31.25\\
Llama3.1 3B it. & 25.99 & 20.32 & 25.97 & 26.90 & 26.26 & 23.23 & 29.41 & 20.94 & 23.67 & 22.92 & 32.82 & 26.09 & 26.21 & 28.05 & 20.12 & 33.16 & 18.18 & 26.74 & 22.70 & 18.75 & 20.65 & 17.02 & 26.73\\
Llama3.2 3B & 20.00 & 24.06 & 19.77 & 25.64 & 17.19 & 20.80 & 25.81 & 26.49 & 24.29 & 20.97 & 14.49 & 22.03 & 25.71 & 20.63 & 19.51 & 31.25 & 24.29 & 26.67 & 17.74 & 21.92 & 18.03 & 18.33 & 21.94\\
Llama3.2 8B it. & 22.38 & 26.82 & 25.00 & \textbf{32.43} & 20.25 & 20.29 & 31.31 & 24.73 & 23.60 & 27.47 & 22.22 & 25.70 & \textbf{30.19} & 25.17 & 20.78 & 32.99 & 28.57 & 21.65 & 23.53 & 24.39 & 28.27 & 22.38 & 25.53\\
Gemma 2 9B  & 17.89 & 19.20 & 16.51 & 16.13 & 17.65 & 10.10 & 14.16 & 20.98 & 14.52 & 11.67 & 14.63 & 13.56 & 16.36 &  9.80 & 10.91 & 18.75 & 17.05 & 16.99 & 12.95 & 15.75 & 13.79 & 10.91 & 18.18 \\
Gemma 2 9B it. & \textbf{28.57} & 31.76 & 27.22 & 28.03 & 25.27 & 20.98 & \textbf{36.88} & \textbf{30.86} & 28.92 & 21.19 & 27.91 & 28.57 & 20.14 & 25.00 & 24.19 & 29.38 & 31.85 & 22.47 & 30.60 & 29.53 & 31.65 & 22.08 & 21.43\\
GPT-3.5-Turbo & 22.22 & 30.77 & 25.13 & 28.71 & 27.72 & 25.64 & 24.56 & 22.01 & 22.66 & 19.08 & 28.31 & 25.89 & 30.15 & 29.51 & 28.16 & 33.00 & 27.20 & 24.03 & \textbf{33.48} & 25.22 & 21.05 & 29.66 & 32.46\\
GPT-4 & 24.80 & \textbf{36.13} & \textbf{30.52} & 30.64 & \textbf{31.91} & \textbf{30.57} & 34.29 & 30.64 & \textbf{35.15} & 31.33 & \textbf{34.01} & \textbf{37.84} & 26.67 & \textbf{37.30} & \textbf{31.30} & \textbf{35.24} & \textbf{38.06} & \textbf{33.91} & 31.54 & \textbf{40.36} & \textbf{36.29} & 30.71 & \textbf{44.21}\\
GPT-4o & 27.16 & 26.77 & 29.03 & 17.43 & 23.19 & 25.00 & 26.05 & 24.80 & 21.62 & 24.79 & 24.19 & 21.65 & 24.27 & 29.18 & 15.51 & 25.78 & 25.70 & 19.24 & 28.57 & 20.82 & 24.10 & 21.71 & 18.25\\
Avg. fs. (w/ GPTs) & 20.82  &  25.25  &  21.06  &  23.94  &  22.21  &  20.15  &  23.99  &  22.47  &  22.82  &  20.78  &  21.34  &  23.98  &  22.89  &  23.05  &  19.07  &  26.13  &  23.33  &  23.56  &  22.85  &  22.83  &  22.49  &  20.35  &  23.78 \\
Avg. fs. (w/o GPTs) & 19.84  &  23.75  &  19.26  &  23.52  &  20.87  &  18.42  &  22.91  &  21.63  &  21.9  &  19.71  &  19.47  &  22.86  &  21.85  &  20.81  &  17.58  &  24.82  &  21.58  &  23.02  &  20.76  &  21.34  &  21.33  &  18.59  &  21.81\\
\midrule
\multicolumn{10}{l}{\textit{Few-shot / Test / Graph\&Node}} \\
CodeLlama 7B & 4.47 & 14.65 &  4.23 &  8.19 &  7.88 &  4.96 &  8.38 & 16.55 &  9.73 & 11.57 &  4.80 & 10.31 & 10.89 &  5.96 &  4.29 &  7.76 &  9.69 & 10.52 & 13.49 & 12.55 &  9.00 &  5.24 &  8.03\\
CodeLlama 7B it. & 10.55 & 20.65 & 12.37 & 17.26 &  9.20 &  9.18 &  5.18 &  9.51 & 18.33 &  9.06 &  7.28 & 10.82 &  7.44 & 15.42 &  3.96 & 18.49 & 11.91 & 11.75 & 10.60 & 11.57 &  6.74 & 16.99 &  9.82\\
CodeLlama 13B & 11.51 &  7.21 &  8.79 & 15.04 &  9.25 & 12.31 &  0.00 & 13.66 & 11.80 &  9.04 & 13.29 & 13.74 & 14.36 &  4.65 &  6.77 & 13.32 & 12.29 & 11.81 & 10.87 & 13.58 & 15.37 &  9.64 &  8.88\\
CodeLlama 13B it. & 6.51 & 14.49 & 10.35 & 12.57 & 18.20 & 11.60 & 21.78 &  7.69 &  5.71 & 11.51 & 10.51 & 19.37 & 16.06 & 10.85 &  8.49 & 12.37 & 12.33 & 20.76 & 12.20 &  6.77 & 19.10 &  8.07 & 12.69\\
Llama3.1 8B &  19.95 & 21.47 & 16.65 & 18.06 & 18.26 & 16.81 & 26.64 & 15.84 & 20.86 & 24.10 & 13.14 & 16.91 & 19.64 & 23.26 & 20.96 & 21.72 & 17.99 & 23.34 & 22.20 & 18.72 & 21.73 & 25.83 & 25.65\\
Llama3.1 3B it. & 17.95 & 14.88 & 17.41 & 21.30 & 20.38 & 17.35 & 21.69 & 14.54 & 17.59 & 16.88 & 23.91 & 19.33 & 17.84 & 20.40 & 15.37 & 27.15 & 14.14 & 18.00 & 17.86 & 12.88 & 14.39 & 10.79 & 19.18\\
Llama3.2 3B & 16.88 & 18.00 & 15.93 & 19.15 & 13.49 & 17.26 & 19.16 & 20.71 & 18.30 & 15.37 & 10.91 & 16.02 & 18.39 & 15.59 & 15.19 & 24.22 & 18.00 & 21.61 & 12.75 & 16.65 & 13.92 & 14.76 & 17.09\\
Llama3.2 8B it. & 15.82 & 21.37 & 18.92 & 24.06 & 15.80 & 14.99 & 23.13 & 20.52 & 19.16 & 19.45 & 17.97 & 19.42 & 24.19 & 17.64 & 15.21 & 27.34 & 21.59 & 17.44 & 17.15 & 20.01 & 21.49 & 17.76 & 21.03 \\
Gemma 2 9B  & 13.43 & 15.02 & 10.75 & 11.49 & 14.05 &  8.92 & 10.34 & 17.31 &  9.68 &  7.89 & 12.20 &  9.07 & 11.52 &  8.26 &  6.69 & 14.59 & 12.28 & 13.76 & 10.90 &  9.75 &  7.76 &  6.65 & 12.70\\
Gemma 2 9B it. & 22.77 & 26.10 & 22.33 & 22.88 & 20.86 & 18.20 & \textbf{29.83} & 26.41 & 24.31 & 17.50 & 24.35 & 22.83 & 17.08 & 21.15 & 19.05 & 26.35 & 25.48 & 18.69 & 25.49 & 23.34 & 24.11 & 18.58 & 16.51\\
GPT-3.5-Turbo & 18.83 & 27.30 & 20.52 & 23.22 & 22.47 & 20.87 & 19.58 & 19.16 & 18.93 & 15.95 & 24.02 & 20.85 & \textbf{24.40} & 23.77 & 22.50 & 29.52 & 21.41 & 21.02 & \textbf{29.27} & 20.33 & 17.52 & 25.73 & 27.94\\
GPT-4 & 21.06 & \textbf{30.99} & \textbf{25.18} & \textbf{25.85} & \textbf{27.61} & \textbf{26.72} & 28.86 & \textbf{27.51} & \textbf{29.82} & \textbf{26.32} & \textbf{30.39} & \textbf{33.28} & 22.82 & \textbf{32.41} & \textbf{25.64} & \textbf{31.28} & \textbf{31.17} & \textbf{30.29} & 27.72 & \textbf{32.77} & \textbf{30.10} & \textbf{27.15} & \textbf{38.74}\\
GPT-4o & \textbf{23.30} & 23.41 & 23.88 & 14.96 & 19.01 & 20.85 & 22.00 & 22.30 & 17.58 & 21.07 & 21.36 & 18.66 & 20.69 & 25.15 & 13.53 & 23.87 & 20.46 & 17.39 & 24.33 & 17.32 & 19.30 & 17.99 & 15.98\\
Avg. fs. (w/ GPTs) & 16.1  &  20.25  &  16.18  &  18.76  &  17.35  &  15.97  &  18.76  &  18.31  &  17.76  &  15.79  &  17.53  &  18.35  &  17.52  &  17.86  &  14.33  &  21.55  &  17.88  &  18.84  &  18.66  &  17.23  &  17.24  &  16.28  &  18.83\\
Avg. fs. (w/o GPTs) & 14.86  &  18.5  &  14.43  &  18.11  &  15.93  &  14.26  &  17.58  &  17.14  &  16.67  &  14.45  &  15.6  &  16.87  &  16.25  &  15.55  &  12.78  &  19.88  &  16.26  &  17.82  &  16.55  &  15.67  &  15.97  &  14.45  &  16.64\\
\midrule
\multicolumn{10}{l}{\textit{Few-shot / Test / Graph\&Node\&Edge}} \\
CodeLlama 7B & 0.00 &  3.02 &  1.68 &  3.85 &  3.27 &  1.33 &  3.09 &  7.23 &  4.03 &  6.85 &  1.15 &  3.16 &  5.70 &  3.12 &  2.35 &  3.89 &  4.30 &  3.61 &  6.75 &  6.35 &  2.85 &  3.70 &  1.96\\
CodeLlama 7B it. & 6.90 & 10.29 &  6.69 & 10.13 &  6.33 &  1.84 &  3.26 &  3.48 &  8.94 &  2.17 &  6.08 &  7.87 &  3.78 &  7.74 &  2.87 &  8.14 &  6.83 &  3.86 &  4.52 &  3.44 &  4.74 & 11.24 &  5.65\\
CodeLlama 13B & 6.20 &  3.88 &  4.06 &  7.99 &  4.85 &  5.02 &  0.00 &  4.43 &  6.79 &  5.40 &  3.99 &  5.12 &  7.49 &  3.21 &  3.95 &  5.04 &  6.06 &  4.07 &  6.02 &  5.76 & 10.00 &  5.22 &  3.28\\
CodeLlama 13B it. & 4.97 &  6.38 &  6.39 &  7.30 & 11.96 &  5.45 & 11.61 &  3.29 &  2.99 &  3.40 &  4.76 &  8.83 &  7.79 &  6.34 &  4.34 &  7.34 &  6.36 & 11.00 &  6.32 &  1.74 & 10.95 &  5.91 &  7.50\\
Llama3.1 8B & 12.46 & 10.70 &  9.92 & 10.81 & 10.36 &  7.03 & 14.26 &  7.35 & 10.35 & 13.61 &  7.02 & 12.11 & 10.65 & 13.93 & 14.19 & 12.29 & 11.30 & 12.08 & 12.57 &  9.85 & 10.91 & 11.58 & 14.33\\
Llama3.1 3B it. & 10.62 &  8.58 &  9.75 & 13.69 & 11.84 &  8.95 & 10.95 &  7.75 &  8.91 &  9.00 & 13.04 & 11.31 & 13.35 & 11.71 &  8.49 & 18.06 &  9.28 & 11.20 &  9.24 &  5.91 &  6.65 &  6.87 & 11.94\\
Llama3.2 3B & 10.01 &  8.59 &  7.68 & 11.15 &  6.93 &  9.08 & 10.44 &  9.45 &  9.83 & 10.67 &  5.09 & 11.70 & 11.53 &  8.17 &  5.68 & 10.66 &  9.84 &  7.52 &  4.26 &  8.59 &  5.49 &  6.93 &  9.00\\
Llama3.2 8B it. & 6.57 & 11.64 & 10.94 & 13.27 &  9.44 & 10.12 & 11.39 & 12.27 & 10.94 & 11.21 & 10.00 & 10.47 & 12.57 & 10.09 &  7.56 & 17.40 & 13.34 & 11.56 & 10.28 & 12.17 & 12.60 &  9.75 & 13.36\\
Gemma 2 9B  & 8.00 &  5.85 &  5.32 &  7.44 &  2.00 &  6.37 &  5.58 & 11.31 &  3.33 &  4.43 &  3.73 &  2.50 &  5.08 &  2.41 &  2.22 &  7.81 &  5.87 &  9.10 &  6.24 &  3.99 &  3.61 &  1.82 &  4.54\\
Gemma 2 9B it. & 13.30 & 15.57 & 11.49 & 14.58 &  9.72 &  9.12 & \textbf{19.61} & 11.76 & 16.83 &  8.36 & 15.17 & 14.01 & 10.57 & 11.83 & 10.56 & 14.89 & 13.53 & 10.44 & 13.19 & 11.79 & 13.60 & 11.36 &  9.00\\
GPT-3.5-Turbo & 12.96 & 17.58 & 11.32 & 15.90 & 15.38 & 12.02 &  9.96 & 12.41 & 12.51 & 10.99 & 13.50 & 10.20 & \textbf{14.43} & 14.40 & 13.58 & 18.46 & 10.18 & 14.76 & 19.58 & 12.87 & 11.49 & 11.87 & 16.30\\
GPT-4 & 14.85 & \textbf{21.14} & \textbf{17.63} & \textbf{19.07} & \textbf{20.53} & \textbf{17.71} & 18.36 & \textbf{19.53} & \textbf{21.16} & \textbf{18.43} & \textbf{20.39} & \textbf{24.32} & 13.91 & \textbf{22.60} & \textbf{16.63} & \textbf{21.06} & \textbf{21.10} & \textbf{22.27} & \textbf{20.47} & \textbf{23.06} & \textbf{21.21} & \textbf{17.99} & \textbf{28.41}\\
GPT-4o & \textbf{15.14} & 17.03 & 14.62 &  9.79 & 12.48 & 14.73 & 14.85 & 16.48 & 11.92 & 13.58 & 12.62 & 12.55 & 14.27 & 17.28 &  9.47 & 16.11 & 12.39 & 12.59 & 17.09 &  9.09 & 12.16 & 12.65 & 11.13\\
Avg. fs. (w/ GPTs) & 9.66  &  11.22  &  8.97  &  11.69  &  10.22  &  8.45  &  10.75  &  10.07  &  10.24  &  9.11  &  9.63  &  10.75  &  10.13  &  10.43  &  8.01  &  12.53  &  10.35  &  10.81  &  10.69  &  9.16  &  9.78  &  9.23  &  10.97\\
Avg. fs. (w/o GPTs) & 8.49  &  9.37  &  7.58  &  10.88  &  8.75  &  6.86  &  9.85  &  8.55  &  9.01  &  7.8  &  8.16  &  9.51  &  9.11  &  8.51  &  6.7  &  11.02  &  9.3  &  9.38  &  8.6  &  7.7  &  8.49  &  8.0  &  9.06\\

\bottomrule
\end{tabular}
    }
    \caption{Scores of \texttt{Graph}, \texttt{Graph\&Node}, and \texttt{Graph\&Node\&Edge} evaluations by 23 languages in EUR-LEX  with \textit{few-shot} models.
    }
    \label{table:scores_by_language_fewshot}
\end{table*}

\begin{table*}[p]
\setlength{\tabcolsep}{3pt}
  \centering
  \footnotesize
  \resizebox{\textwidth}{!}{
    \begin{tabular}{lcccccccccccccccccccccccc}
\toprule
            \textbf{Model} & \textbf{BG} & \textbf{ES} & \textbf{CS} & \textbf{DA} & \textbf{DE} & \textbf{ET} & \textbf{EL} & \textbf{EN} & \textbf{FR} & \textbf{HR} & \textbf{IT} & \textbf{LV} & \textbf{LT} & \textbf{HU} & \textbf{MT} & \textbf{NL} & \textbf{PL} & \textbf{PT} & \textbf{RO} & \textbf{SK} & \textbf{SL} & \textbf{FI} & \textbf{SV} \\
\midrule
\multicolumn{11}{l}{\textit{Finetuning / Test / Graph}} \\
CodeLlama 7B & 29.93 & 33.33 & 28.78 & 33.55 & 27.59 & 31.72 & 23.53 & 37.87 & 25.97 & 38.22 & 30.30 & 30.97 & 35.53 & 26.57 & 34.90 & 25.97 & 32.89 & 34.78 & 23.60 & 29.37 & 25.53 & 28.19 & 31.17\\
CodeLlama 7B it. & 37.27 & 35.58 & 30.23 & 30.60 & 37.65 & 31.45 & 22.38 & 36.78 & 33.13 & 40.24 & 27.50 & 38.10 & 37.50 & 35.58 & 37.43 & 26.99 & 37.57 & 31.52 & 28.40 & 34.57 & 34.12 & 26.99 & 36.25\\
CodeLlama 13B & 35.62 & 37.09 & 30.67 & 38.71 & 37.27 & 32.70 & 33.33 & 41.03 & 34.84 & 35.00 & 32.43 & 40.48 & 35.06 & 27.27 & 36.94 & 26.75 & 33.12 & 29.33 & 42.47 & 31.58 & 36.13 & 31.79 & 32.10\\
CodeLlama 13B it. & 31.06 & 39.26 & 40.48 & 36.36 & 33.77 & 30.67 & 24.82 & 36.78 & 34.44 & 35.93 & 29.11 & 36.69 & 44.58 & 32.34 & 34.78 & 35.80 & 41.21 & 30.49 & 36.71 & 42.29 & 35.58 & 32.94 & 39.33\\
Llama3.1 8B & 33.75 & 31.79 & 22.07 & 35.53 & 27.50 & 27.21 & 27.97 & 31.06 & 24.11 & 41.29 & 37.74 & 37.18 & 35.67 & 23.45 & 33.99 & 28.38 & 28.00 & 28.77 & 22.78 & 33.99 & 29.93 & 26.75 & 20.69\\
Llama3.1 3B it. & 41.03 & 29.33 & 29.93 & 24.52 & 26.67 & 32.50 & 31.08 & 34.15 & 26.92 & 30.26 & 22.22 & 35.37 & 30.57 & 24.16 & 28.40 & 21.62 & 27.81 & 31.17 & 27.50 & 30.26 & 33.55 & 30.67 & 29.53\\
Llama3.2 3B & 30.59 & 32.18 & 35.71 & 36.14 & 29.09 & 36.57 & 42.11 & 35.43 & 26.37 & 30.77 & 27.54 & 37.29 & 40.24 & 32.93 & 34.32 & 32.61 & 29.94 & 32.56 & 36.99 & 32.10 & 30.30 & 32.40 & 33.51\\
Llama3.2 8B it. & 24.04 & 20.86 & 29.51 & 31.58 & 28.74 & 30.77 & 23.46 & 30.67 & 29.45 & 41.57 & 22.73 & 34.44 & 38.42 & 24.39 & 29.35 & 29.83 & 27.81 & 28.74 & 33.33 & 37.36 & 34.83 & 31.35 & 33.96\\
Gemma 2 9B  & 41.86 & \textbf{44.69} & \textbf{46.34} & 47.13 & \textbf{41.92} & 37.93 & 38.64 & \textbf{44.83} & 30.23 & \textbf{50.00} & \textbf{39.33} & 42.70 & 40.94 & \textbf{49.70} & \textbf{41.34} & 40.68 & \textbf{50.29} & \textbf{45.16} & \textbf{47.95} & \textbf{47.06} & 40.64 & \textbf{44.58} & 44.31\\
Gemma 2 9B it. & \textbf{47.13} & 38.51 & 42.42 & \textbf{47.85} & 41.42 & 38.51 & \textbf{44.71} & 43.87 & \textbf{43.93} & 41.77 & 38.99 & \textbf{43.64} & \textbf{45.51} & 38.60 & 32.94 & \textbf{42.17} & 45.68 & 41.92 & 41.98 & 36.59 & \textbf{44.30} & 43.11 & \textbf{47.62}\\
Avg. of all ft. & 36.31  &  34.9  &  34.6  &  37.53  &  35.69  &  \textbf{38.52}  &  28.2  &  37.08  &  28.4  &  38.71  &  35.44  &  36.42  &  37.94  &  31.74  &  30.6  &  27.23  &  33.55  &  29.28  &  29.27  &  34.84  &  35.43  &  27.97  &  30.0\\

\midrule
\multicolumn{11}{l}{\textit{Finetuning / Test / Graph\&Node}} \\
CodeLlama 7B & 21.28 & 27.17 & 21.52 & 25.26 & 20.33 & 22.56 & 17.04 & 31.29 & 18.23 & 29.44 & 23.14 & 23.12 & 26.45 & 19.04 & 26.64 & 19.56 & 24.96 & 27.01 & 19.00 & 19.47 & 18.84 & 20.31 & 25.17\\
CodeLlama 7B it. & 26.52 & 29.07 & 23.18 & 25.38 & 29.25 & 22.78 & 16.09 & 30.40 & 26.42 & 31.85 & 22.05 & 28.34 & 27.88 & 26.60 & 27.31 & 21.66 & 26.52 & 26.09 & 23.89 & 27.31 & 25.94 & 20.44 & 27.89\\
CodeLlama 13B & 26.33 & 30.52 & 24.32 & 29.04 & 27.33 & 23.90 & 22.88 & 33.87 & 27.74 & 26.07 & 24.08 & 28.97 & 25.05 & 18.78 & 26.42 & 21.54 & 23.46 & 22.71 & 35.20 & 22.91 & 28.17 & 22.64 & 24.54\\
CodeLlama 13B it. & 23.57 & 31.71 & 32.40 & 30.21 & 26.07 & 22.67 & 17.95 & 30.77 & 27.17 & 27.72 & 21.07 & 28.93 & 31.87 & 22.91 & 26.60 & 28.58 & 32.49 & 24.96 & 30.24 & 32.49 & 28.13 & 25.42 & 31.29\\
Llama3.1 8B & 18.99 & 22.06 & 13.76 & 23.30 & 17.93 & 17.57 & 18.12 & 23.40 & 16.16 & 25.72 & 24.98 & 23.53 & 23.77 & 16.56 & 23.11 & 18.96 & 18.63 & 19.75 & 15.41 & 23.37 & 20.29 & 17.08 & 12.74\\
Llama3.1 3B it. & 26.68 & 20.05 & 19.56 & 18.36 & 20.13 & 22.06 & 20.39 & 26.75 & 18.11 & 20.27 & 15.42 & 23.91 & 21.08 & 16.60 & 19.73 & 14.74 & 18.34 & 21.48 & 19.65 & 20.93 & 21.67 & 19.89 & 20.45\\
Llama3.2 3B & 21.75 & 25.01 & 25.27 & 26.81 & 20.49 & 25.98 & 28.95 & 29.00 & 18.74 & 22.66 & 19.36 & 26.49 & 28.97 & 22.71 & 24.40 & 24.29 & 22.07 & 24.23 & 28.33 & 22.64 & 22.52 & 23.68 & 24.06\\
Llama3.2 8B it. & 16.04 & 15.71 & 20.47 & 23.16 & 20.44 & 21.44 & 16.81 & 24.92 & 19.54 & 29.63 & 16.04 & 23.85 & 26.68 & 17.21 & 21.26 & 22.56 & 17.77 & 20.43 & 23.57 & 26.12 & 22.82 & 22.91 & 25.02\\
Gemma 2 9B  & 35.84 & \textbf{39.30} & \textbf{39.52} & \textbf{40.36} & \textbf{35.99} & \textbf{30.88} & 31.56 & \textbf{39.37} & 25.36 & \textbf{39.15} & 32.65 & 35.50 & 32.74 & \textbf{40.44} & \textbf{35.84} & 33.84 & \textbf{41.89} & \textbf{38.60} & \textbf{42.59} & \textbf{40.51} & 33.32 & \textbf{37.28} & 36.72\\
Gemma 2 9B it. & \textbf{38.70} & 32.43 & 34.57 & 38.47 & 33.45 & 27.28 & \textbf{33.46} & 37.36 & \textbf{34.58} & 33.65 & \textbf{33.95} & \textbf{35.84} & \textbf{35.03} & 31.80 & 25.65 & \textbf{34.49} & 35.94 & 36.26 & 36.12 & 29.92 & \textbf{35.27} & 35.48 & \textbf{38.84}\\
Avg. of all ft. & 26.82  &  27.47  &  25.98  &  28.6  &  26.97  &  26.44  &  19.62  &  29.8  &  20.61  &  27.52  &  25.5  &  25.55  &  27.32  &  21.69  &  21.95  &  20.28  &  24.88  &  22.18  &  22.7  &  25.28  &  25.84  &  20.25  &  22.29\\

\midrule
\multicolumn{11}{l}{\textit{Finetuning / Test / Graph\&Node\&Edge}} \\
CodeLlama 7B & 14.83 & 19.29 & 15.09 & 18.38 & 14.93 & 16.34 & 13.10 & 23.14 & 13.30 & 22.20 & 14.33 & 16.40 & 18.48 & 14.42 & 18.28 & 13.37 & 17.82 & 20.31 & 13.72 & 12.41 & 12.65 & 13.28 & 17.55\\
CodeLlama 7B it. & 18.09 & 21.38 & 16.39 & 18.46 & 20.97 & 16.77 & 12.10 & 24.87 & 20.06 & 21.54 & 17.01 & 20.69 & 18.84 & 19.17 & 18.71 & 14.42 & 18.32 & 19.54 & 17.39 & 20.30 & 17.95 & 14.20 & 21.30\\
CodeLlama 13B & 18.20 & 21.93 & 16.02 & 20.54 & 18.84 & 15.03 & 15.72 & 24.08 & 20.00 & 16.99 & 16.44 & 16.75 & 18.01 & 13.45 & 16.56 & 14.80 & 16.92 & 16.64 & 24.63 & 12.79 & 19.35 & 15.95 & 17.66\\
CodeLlama 13B it. & 19.45 & 22.43 & 22.36 & 21.65 & 18.98 & 13.61 & 12.76 & 24.73 & 18.66 & 21.50 & 15.10 & 20.08 & 20.67 & 15.66 & 18.07 & 19.11 & 23.81 & 18.13 & 19.53 & 23.49 & 18.72 & 17.69 & 22.96\\
Llama3.1 8B & 11.00 & 15.08 & 10.62 & 16.32 & 12.95 & 11.50 & 12.32 & 18.07 & 11.94 & 15.42 & 17.07 & 14.54 & 15.12 & 11.09 & 16.48 & 10.77 & 13.47 & 13.90 &  8.97 & 15.09 & 13.53 &  9.58 &  9.16\\
Llama3.1 3B it. & 16.76 & 10.93 & 13.38 & 13.86 & 14.57 & 15.49 & 16.79 & 20.36 & 12.40 & 12.40 &  8.35 & 14.93 & 13.41 &  9.12 & 11.11 &  8.81 & 13.84 & 14.36 & 13.35 & 13.99 & 11.77 & 13.20 & 14.79\\
Llama3.2 3B & 16.68 & 18.57 & 18.65 & 19.94 & 15.08 & 18.69 & 21.78 & 22.81 & 14.33 & 15.02 & 13.21 & 18.67 & 19.27 & 15.31 & 16.59 & 18.76 & 16.69 & 16.45 & 20.83 & 15.71 & 14.94 & 17.70 & 17.57\\
Llama3.2 8B it. & 11.32 & 12.62 & 14.19 & 15.37 & 13.53 & 14.62 & 10.73 & 20.60 & 13.08 & 19.75 & 11.38 & 15.33 & 17.70 & 12.62 & 15.08 & 14.63 & 10.86 & 15.19 & 15.49 & 16.55 & 14.95 & 14.87 & 18.02\\
Gemma 2 9B  & 27.14 & \textbf{29.70} & \textbf{26.62} & \textbf{31.74} & \textbf{27.95} & 21.01 & \textbf{24.89} & \textbf{32.29} & 19.77 & \textbf{28.83} & 23.09 & 25.83 & 23.11 & \textbf{30.80} & \textbf{27.63} & 25.04 & \textbf{33.68} & \textbf{29.35} & \textbf{33.13} & \textbf{30.44} & 24.77 & \textbf{28.79} & 27.55\\
Gemma 2 9B it. & \textbf{28.05} & 24.61 & 25.58 & 28.85 & 23.58 & \textbf{22.44} & 24.39 & 31.28 & \textbf{27.55} & 25.58 & \textbf{25.51} & \textbf{27.46} & \textbf{24.83} & 23.74 & 18.52 & \textbf{25.72} & 26.32 & 29.08 & 28.25 & 21.73 & \textbf{26.26} & 26.57 & \textbf{31.51}\\
Avg. of all ft. & 19.28  &  19.89  &  18.74  &  21.46  &  17.47  &  18.74  &  14.46  &  21.43  &  14.79  &  19.16  &  16.23  &  17.24  &  19.16  &  15.69  &  14.79  &  13.96  &  18.09  &  16.3  &  15.89  &  17.4  &  17.68  &  13.99  &  16.53\\

\bottomrule
\end{tabular}
    }
    \caption{Scores of \texttt{Graph}, \texttt{Graph\&Node}, and \texttt{Graph\&Node\&Edge} evaluations by 23 languages in EUR-LEX with \textit{finetuned} models.
    }
    \label{table:scores_by_language_finetuned}
\end{table*}

%% file: tables/main_result_appendix.tex
\begin{table*}[t]
\centering
\footnotesize
\begin{tabular}{lccccccccc}
\toprule
    &\multicolumn{3}{c}{\textbf{Graph-based}}  & \multicolumn{2}{c}{\textbf{Valid Graph}} & \multicolumn{4}{c}{\textbf{Legal Content}} \\
    \cmidrule(lr){2-4} \cmidrule(lr){5-6} \cmidrule(lr){7-10}
\textbf{Model} &\textbf{G} & \textbf{G-N} & \textbf{G-N-E} & \textbf{Top1} & \textbf{Top10} & \textbf{Entity} & \textbf{R \& T} & \textbf{Source} & \textbf{Statement} \\
\midrule
\multicolumn{7}{l}{\textit{Few-shot result of Validation split}} \\
CodeLlama 7B & 14.46 & 10.14 & 4.73 & 18.35 & 86.96 & 47.03 & 6.51 & 12.44 & 1.78 \\
CodeLlama 7B it. & 16.70 & 12.21 & 6.64 & 43.30 & 91.91 & 52.18 & 8.91 & 17.67 & 1.71 \\
CodeLlama 13B & 14.99 & 10.53 & 5.17 & 18.09 & 84.96 & 47.58 & 7.14 & 14.93 & 2.76\\
CodeLlama 13B it. \hspace{-1em} & 18.02 & 13.33 & 6.82 & 38.26  & 89.74 &  54.30 & 8.98 & 18.65 & 3.75 \\
Llama3.1 8B & 24.87 & 19.05 & 10.07 & 36.17 & 87.04 & 60.77 & 12.74 & 23.86 & 4.17\\
Llama3.1 8B it. \hspace{-1em} & 26.88 & 20.00 & 11.42 & 26.96 & 88.00 & 60.86 & 15.02 & 29.35 & 3.77\\
Llama3.2 3B & 21.04 & 16.00 & 7.92 & 31.91 & 85.83 & 54.39 & 10.25 & 21.69 & 2.89\\
Llama3.2 3B it. \hspace{-1em} &  26.45 & 20.21 & 11.45 & 61.04 & 93.83 &  53.56 & 14.78 & 20.63 & 3.69 \\
Gemma2 9B  & 12.22 & 8.80 & 4.08 & 42.61 & 96.35 & 52.40 & 5.38 & 14.83 & 2.84\\
Gemma2 9B it. & 27.66 & 21.92 & 13.37 & 71.04 & 96.87 & 69.11 & 16.81 & 30.67 & 3.36\\
GPT-3.5-Turbo & 31.45 & 25.62 & 16.41 & 96.87 & \textbf{100.0} & 69.80 & 19.78 & \textbf{29.43} & \textbf{5.48}\\
GPT-4         & \textbf{33.09} & \textbf{27.34} & \textbf{19.15} & \textbf{98.96} & \textbf{100.0} & \textbf{75.31} & \textbf{23.24} & 21.52 & 3.30\\
GPT-4o        & 30.98 & 25.79 & 17.24 & 95.74 & \textbf{100.0} & 71.61 & 20.40 & 40.72 & 4.77\\
\midrule
\multicolumn{7}{l}{\textit{Finetuning result of Validation split}} \\
CodeLlama 7B & 36.38 & 29.66 & 21.57 & 95.39 & \textbf{100.0} & 75.47 & 26.10 & 48.51 & 7.22\\
CodeLlama 7B it. \hspace{-1em}  & 35.86 & 29.40 & 21.91 & 96.87 & 99.91 &  76.07 & 26.22 & 46.18 & 5.92\\
CodeLlama 13B  & 35.94 & 29.06 & 20.05 & 97.48 & \textbf{100.0} & 74.38 & 24.36 & 50.47 & 8.15\\
CodeLlama 13B it.  & 33.81 & 27.63 & 19.91 & 97.57 & \textbf{100.0} & 75.96 & 24.07 & 44.32 & 6.27\\
Llama3.1 8B  & 30.30 & 21.78 & 14.77 & 94.52 & \textbf{100.0} & 64.19 & 20.29 & 31.44 & 3.75\\
Llama3.1 8B it. \hspace{-1em}  & 29.14 & 20.92 & 13.53 & 85.30 & 99.74 &  66.77 & 18.21 & 41.31 & 5.39\\
Llama3.2 3B  & 31.56 & 24.34 & 17.48 & 93.57 & \textbf{100.0} & 71.02 & 22.31 & 47.09 & 5.65\\
Llama3.2 3B it. \hspace{-1em}  & 31.70 & 24.06 & 16.13 & 90.00 & 99.83 &  68.65 & 20.94 & 43.29 & 4.85\\
Gemma2 9B   & 43.62 & 37.32 & \textbf{28.37} & \textbf{99.13} & \textbf{100.0} & \textbf{79.31} & \textbf{32.65} & \textbf{58.96} & 8.99 \\
Gemma2 9B it.  & \textbf{43.88} & \textbf{36.57} & 27.36 & 98.17 & \textbf{100.0} & 77.72 & 32.41 & 53.20 & \textbf{10.51}\\
\bottomrule
\end{tabular}
\caption{
Scores of the legal text visualization. 
\textbf{G}, \textbf{G-N} and \textbf{G-N-E} denote \texttt{Graph}, \texttt{Graph\&Node} and  \texttt{Graph\&Node\&Edge} respectively. Valid Graph Ratio is success rate of creating valid graphs in top-1 and top-10 generated results.
The highest scores of each column are in bold.
}
\label{table:main_result_appendix}
\end{table*}

%% file: tables/prompt_table.tex
\begin{table*}[h]
    \centering
    \footnotesize
        \begin{tabularx}{\linewidth}{p{30mm}X}
          \toprule
              \textbf{Method} & \textbf{Prompt} \\
          \midrule
         \cellcolor{purple} Prompt used for train and generation & Using the DOT language of Graphviz, draw a graph to explain legal entity nodes, legal relationships, legal statements and legal basis of them from given text, written in \{language\} text.
Use “shape=trapezium” to represent a legally effective material and use “shape=doubleoctagon” to represent a legal entity in Graphviz code with \{language\}.
At any time, reply only with the graphviz code.
 \\
         \midrule
         \cellcolor{purple} Prompt for extraction & From legal text below of \{language\} language, extract the same meaning word or sentence as given English word to language. Please output only extracted result. Legal text: \{legal text\} Word or sentence to extract: \\
         \midrule
         \cellcolor{purple} Prompt for translation & Translate below words or text from English to \{language\}   Text: \\
          \bottomrule
    \end{tabularx}
    \caption{
      The prompts used in the experiment and data processing. \{legal text\} and \{language\} indicate the place to insert.
    }
    \label{tab:app-prompt}
\end{table*}

%% file: acl_latex.bbl
\begin{thebibliography}{52}
\providecommand{\natexlab}[1]{#1}

\bibitem[{Angelidis et~al.(2018)Angelidis, Chalkidis, and
  Koubarakis}]{Angelidis2018NamedER}
Iosif Angelidis, Ilias Chalkidis, and Manolis Koubarakis. 2018.
\newblock \href {https://api.semanticscholar.org/CorpusID:55699546} {Named
  entity recognition, linking and generation for greek legislation}.
\newblock In \emph{International Conference on Legal Knowledge and Information
  Systems}.

\bibitem[{Aumiller et~al.(2021)Aumiller, Almasian, Lackner, and
  Gertz}]{10.1145/3462757.3466085}
Dennis Aumiller, Satya Almasian, Sebastian Lackner, and Michael Gertz. 2021.
\newblock \href {https://doi.org/10.1145/3462757.3466085} {Structural text
  segmentation of legal documents}.
\newblock In \emph{Proceedings of the Eighteenth International Conference on
  Artificial Intelligence and Law}, ICAIL '21, page 2–11, New York, NY, USA.
  Association for Computing Machinery.

\bibitem[{Aumiller et~al.(2022)Aumiller, Chouhan, and
  Gertz}]{aumiller-etal-2022-eur}
Dennis Aumiller, Ashish Chouhan, and Michael Gertz. 2022.
\newblock \href {https://doi.org/10.18653/v1/2022.emnlp-main.519}
  {{EUR}-lex-sum: A multi- and cross-lingual dataset for long-form
  summarization in the legal domain}.
\newblock In \emph{Proceedings of the 2022 Conference on Empirical Methods in
  Natural Language Processing}, pages 7626--7639, Abu Dhabi, United Arab
  Emirates. Association for Computational Linguistics.

\bibitem[{Barale et~al.(2023)Barale, Klaisoongnoen, Minervini, Rovatsos, and
  Bhuta}]{barale-etal-2023-asylex}
Claire Barale, Mark Klaisoongnoen, Pasquale Minervini, Michael Rovatsos, and
  Nehal Bhuta. 2023.
\newblock \href {https://doi.org/10.18653/v1/2023.nllp-1.24} {{A}sy{L}ex: A
  dataset for legal language processing of refugee claims}.
\newblock In \emph{Proceedings of the Natural Legal Language Processing
  Workshop 2023}, pages 244--257, Singapore. Association for Computational
  Linguistics.

\bibitem[{Belouadi et~al.(2024)Belouadi, Lauscher, and Eger}]{automatikz}
Jonas Belouadi, Anne Lauscher, and Steffen Eger. 2024.
\newblock Automatikz: Text-guided synthesis of scientific vector graphics with
  tikz.
\newblock In \emph{International Conference on Learning Representations
  (ICLR)}.

\bibitem[{Brown et~al.(2020)Brown, Mann, Ryder, Subbiah, Kaplan, Dhariwal,
  Neelakantan, Shyam, Sastry, Askell, Agarwal, Herbert-Voss, Krueger, Henighan,
  Child, Ramesh, Ziegler, Wu, Winter, Hesse, Chen, Sigler, Litwin, Gray, Chess,
  Clark, Berner, McCandlish, Radford, Sutskever, and
  Amodei}]{NEURIPS2020_1457c0d6}
Tom Brown, Benjamin Mann, Nick Ryder, Melanie Subbiah, Jared~D Kaplan, Prafulla
  Dhariwal, Arvind Neelakantan, Pranav Shyam, Girish Sastry, Amanda Askell,
  Sandhini Agarwal, Ariel Herbert-Voss, Gretchen Krueger, Tom Henighan, Rewon
  Child, Aditya Ramesh, Daniel Ziegler, Jeffrey Wu, Clemens Winter, Chris
  Hesse, Mark Chen, Eric Sigler, Mateusz Litwin, Scott Gray, Benjamin Chess,
  Jack Clark, Christopher Berner, Sam McCandlish, Alec Radford, Ilya Sutskever,
  and Dario Amodei. 2020.
\newblock \href
  {https://proceedings.neurips.cc/paper_files/paper/2020/file/1457c0d6bfcb4967418bfb8ac142f64a-Paper.pdf}
  {Language models are few-shot learners}.
\newblock In \emph{Advances in Neural Information Processing Systems},
  volume~33, pages 1877--1901. Curran Associates, Inc.

\bibitem[{Bubeck et~al.(2023)Bubeck, Chandrasekaran, Eldan, Gehrke, Horvitz,
  Kamar, Lee, Lee, Li, Lundberg, Nori, Palangi, Ribeiro, and
  Zhang}]{SparksofAGI}
S{\'e}bastien Bubeck, Varun Chandrasekaran, Ronen Eldan, John~A. Gehrke, Eric
  Horvitz, Ece Kamar, Peter Lee, Yin~Tat Lee, Yuan-Fang Li, Scott~M. Lundberg,
  Harsha Nori, Hamid Palangi, Marco~Tulio Ribeiro, and Yi~Zhang. 2023.
\newblock \href {https://api.semanticscholar.org/CorpusID:257663729} {Sparks of
  artificial general intelligence: Early experiments with gpt-4}.
\newblock \emph{ArXiv}, abs/2303.12712.

\bibitem[{Chalkidis et~al.(2019)Chalkidis, Fergadiotis, Malakasiotis, Aletras,
  and Androutsopoulos}]{chalkidis-etal-2019-extreme}
Ilias Chalkidis, Emmanouil Fergadiotis, Prodromos Malakasiotis, Nikolaos
  Aletras, and Ion Androutsopoulos. 2019.
\newblock \href {https://doi.org/10.18653/v1/W19-2209} {Extreme multi-label
  legal text classification: A case study in {EU} legislation}.
\newblock In \emph{Proceedings of the Natural Legal Language Processing
  Workshop 2019}, pages 78--87, Minneapolis, Minnesota. Association for
  Computational Linguistics.

\bibitem[{Chalkidis et~al.(2021{\natexlab{a}})Chalkidis, Fergadiotis, and
  Androutsopoulos}]{chalkidis-etal-2021-multieurlex}
Ilias Chalkidis, Manos Fergadiotis, and Ion Androutsopoulos.
  2021{\natexlab{a}}.
\newblock \href {https://doi.org/10.18653/v1/2021.emnlp-main.559}
  {{M}ulti{EURLEX} - a multi-lingual and multi-label legal document
  classification dataset for zero-shot cross-lingual transfer}.
\newblock In \emph{Proceedings of the 2021 Conference on Empirical Methods in
  Natural Language Processing}, pages 6974--6996, Online and Punta Cana,
  Dominican Republic. Association for Computational Linguistics.

\bibitem[{Chalkidis et~al.(2021{\natexlab{b}})Chalkidis, Fergadiotis,
  Tsarapatsanis, Aletras, Androutsopoulos, and
  Malakasiotis}]{chalkidis-etal-2021-paragraph}
Ilias Chalkidis, Manos Fergadiotis, Dimitrios Tsarapatsanis, Nikolaos Aletras,
  Ion Androutsopoulos, and Prodromos Malakasiotis. 2021{\natexlab{b}}.
\newblock \href {https://doi.org/10.18653/v1/2021.naacl-main.22}
  {Paragraph-level rationale extraction through regularization: A case study on
  {E}uropean court of human rights cases}.
\newblock In \emph{Proceedings of the 2021 Conference of the North American
  Chapter of the Association for Computational Linguistics: Human Language
  Technologies}, pages 226--241, Online. Association for Computational
  Linguistics.

\bibitem[{Chalkidis et~al.(2023)Chalkidis, Garneau, Goanta, Katz, and
  S{\o}gaard}]{chalkidis-etal-2023-lexfiles}
Ilias Chalkidis, Nicolas Garneau, Catalina Goanta, Daniel Katz, and Anders
  S{\o}gaard. 2023.
\newblock \href {https://doi.org/10.18653/v1/2023.acl-long.865} {{L}e{XF}iles
  and {L}egal{LAMA}: Facilitating {E}nglish multinational legal language model
  development}.
\newblock In \emph{Proceedings of the 61st Annual Meeting of the Association
  for Computational Linguistics (Volume 1: Long Papers)}, pages 15513--15535,
  Toronto, Canada. Association for Computational Linguistics.

\bibitem[{Chalkidis et~al.(2022{\natexlab{a}})Chalkidis, Jana, Hartung,
  Bommarito, Androutsopoulos, Katz, and Aletras}]{chalkidis-etal-2022-lexglue}
Ilias Chalkidis, Abhik Jana, Dirk Hartung, Michael Bommarito, Ion
  Androutsopoulos, Daniel Katz, and Nikolaos Aletras. 2022{\natexlab{a}}.
\newblock \href {https://doi.org/10.18653/v1/2022.acl-long.297} {{L}ex{GLUE}: A
  benchmark dataset for legal language understanding in {E}nglish}.
\newblock In \emph{Proceedings of the 60th Annual Meeting of the Association
  for Computational Linguistics (Volume 1: Long Papers)}, pages 4310--4330,
  Dublin, Ireland. Association for Computational Linguistics.

\bibitem[{Chalkidis et~al.(2022{\natexlab{b}})Chalkidis, Pasini, Zhang, Tomada,
  Schwemer, and S{\o}gaard}]{chalkidis-etal-2022-fairlex}
Ilias Chalkidis, Tommaso Pasini, Sheng Zhang, Letizia Tomada, Sebastian
  Schwemer, and Anders S{\o}gaard. 2022{\natexlab{b}}.
\newblock \href {https://doi.org/10.18653/v1/2022.acl-long.301} {{F}air{L}ex: A
  multilingual benchmark for evaluating fairness in legal text processing}.
\newblock In \emph{Proceedings of the 60th Annual Meeting of the Association
  for Computational Linguistics (Volume 1: Long Papers)}, pages 4389--4406,
  Dublin, Ireland. Association for Computational Linguistics.

\bibitem[{Chen and Eagel(2017)}]{10.1145/3086512.3086538}
Daniel~L. Chen and Jess Eagel. 2017.
\newblock \href {https://doi.org/10.1145/3086512.3086538} {Can machine learning
  help predict the outcome of asylum adjudications?}
\newblock In \emph{Proceedings of the 16th Edition of the International
  Conference on Articial Intelligence and Law}, ICAIL '17, page 237–240, New
  York, NY, USA. Association for Computing Machinery.

\bibitem[{Chen et~al.(2019)Chen, Cai, Dai, Dai, and
  Ding}]{chen-etal-2019-charge}
Huajie Chen, Deng Cai, Wei Dai, Zehui Dai, and Yadong Ding. 2019.
\newblock \href {https://doi.org/10.18653/v1/D19-1667} {Charge-based prison
  term prediction with deep gating network}.
\newblock In \emph{Proceedings of the 2019 Conference on Empirical Methods in
  Natural Language Processing and the 9th International Joint Conference on
  Natural Language Processing (EMNLP-IJCNLP)}, pages 6362--6367, Hong Kong,
  China. Association for Computational Linguistics.

\bibitem[{Choi et~al.(2021)Choi, Hickman, Monahan, and
  Schwarcz}]{choi2021chatgpt}
Jonathan~H Choi, Kristin~E Hickman, Amy~B Monahan, and Daniel Schwarcz. 2021.
\newblock Chatgpt goes to law school.
\newblock \emph{J. Legal Educ.}, 71:387.

\bibitem[{Christopoulou et~al.(2024)Christopoulou, Zhang, and
  Lampouras}]{christopoulou-etal-2024-text}
Fenia Christopoulou, Guchun Zhang, and Gerasimos Lampouras. 2024.
\newblock \href {https://aclanthology.org/2024.eacl-long.72} {Text-to-code
  generation with modality-relative pre-training}.
\newblock In \emph{Proceedings of the 18th Conference of the European Chapter
  of the Association for Computational Linguistics (Volume 1: Long Papers)},
  pages 1194--1208, St. Julian{'}s, Malta. Association for Computational
  Linguistics.

\bibitem[{de~Gibert~Bonet et~al.(2022)de~Gibert~Bonet, Garc{\'\i}a~Pablos,
  Cuadros, and Melero}]{de-gibert-bonet-etal-2022-spanish}
Ona de~Gibert~Bonet, Aitor Garc{\'\i}a~Pablos, Montse Cuadros, and Maite
  Melero. 2022.
\newblock \href {https://aclanthology.org/2022.lrec-1.400} {{S}panish datasets
  for sensitive entity detection in the legal domain}.
\newblock In \emph{Proceedings of the Thirteenth Language Resources and
  Evaluation Conference}, pages 3751--3760, Marseille, France. European
  Language Resources Association.

\bibitem[{Drawzeski et~al.(2021)Drawzeski, Galassi, Jablonowska, Lagioia,
  Lippi, Micklitz, Sartor, Tagiuri, and Torroni}]{drawzeski-etal-2021-corpus}
Kasper Drawzeski, Andrea Galassi, Agnieszka Jablonowska, Francesca Lagioia,
  Marco Lippi, Hans~Wolfgang Micklitz, Giovanni Sartor, Giacomo Tagiuri, and
  Paolo Torroni. 2021.
\newblock \href {https://doi.org/10.18653/v1/2021.nllp-1.1} {A corpus for
  multilingual analysis of online terms of service}.
\newblock In \emph{Proceedings of the Natural Legal Language Processing
  Workshop 2021}, pages 1--8, Punta Cana, Dominican Republic. Association for
  Computational Linguistics.

\bibitem[{Dubey et~al.(2024)Dubey, Jauhri, Pandey, Kadian, Al-Dahle
  et~al.}]{dubey2024llama3herdmodels}
Abhimanyu Dubey, Abhinav Jauhri, Abhinav Pandey, Abhishek Kadian, Ahmad
  Al-Dahle, et~al. 2024.
\newblock \href {https://arxiv.org/abs/2407.21783} {The llama 3 herd of
  models}.
\newblock \emph{Preprint}, arXiv:2407.21783.

\bibitem[{Dunn et~al.(2017)Dunn, Sagun, \c{S}irin, and
  Chen}]{10.1145/3086512.3086537}
Matt Dunn, Levent Sagun, Hale \c{S}irin, and Daniel Chen. 2017.
\newblock \href {https://doi.org/10.1145/3086512.3086537} {Early predictability
  of asylum court decisions}.
\newblock In \emph{Proceedings of the 16th Edition of the International
  Conference on Articial Intelligence and Law}, ICAIL '17, page 233–236, New
  York, NY, USA. Association for Computing Machinery.

\bibitem[{Elaraby and Litman(2022)}]{elaraby-litman-2022-arglegalsumm}
Mohamed Elaraby and Diane Litman. 2022.
\newblock \href {https://aclanthology.org/2022.coling-1.540}
  {{A}rg{L}egal{S}umm: Improving abstractive summarization of legal documents
  with argument mining}.
\newblock In \emph{Proceedings of the 29th International Conference on
  Computational Linguistics}, pages 6187--6194, Gyeongju, Republic of Korea.
  International Committee on Computational Linguistics.

\bibitem[{Frankenreiter and Nyarko(2022)}]{frankenreiter2022natural}
Jens Frankenreiter and Julian Nyarko. 2022.
\newblock Natural language processing in legal tech.
\newblock \emph{Legal Tech and the Future of Civil Justice}.

\bibitem[{Guha et~al.(2023)Guha, Nyarko, Ho, Ré, Chilton, Narayana,
  Chohlas-Wood, Peters, Waldon, Rockmore, Zambrano, Talisman, Hoque, Surani,
  Fagan, Sarfaty, Dickinson, Porat, Hegland, Wu, Nudell, Niklaus, Nay, Choi,
  Tobia, Hagan, Ma, Livermore, Rasumov-Rahe, Holzenberger, Kolt, Henderson,
  Rehaag, Goel, Gao, Williams, Gandhi, Zur, Iyer, and Li}]{legalbench2023}
Neel Guha, Julian Nyarko, Daniel~E. Ho, Christopher Ré, Adam Chilton, Aditya
  Narayana, Alex Chohlas-Wood, Austin Peters, Brandon Waldon, Daniel~N.
  Rockmore, Diego Zambrano, Dmitry Talisman, Enam Hoque, Faiz Surani, Frank
  Fagan, Galit Sarfaty, Gregory~M. Dickinson, Haggai Porat, Jason Hegland,
  Jessica Wu, Joe Nudell, Joel Niklaus, John Nay, Jonathan~H. Choi, Kevin
  Tobia, Margaret Hagan, Megan Ma, Michael Livermore, Nikon Rasumov-Rahe, Nils
  Holzenberger, Noam Kolt, Peter Henderson, Sean Rehaag, Sharad Goel, Shang
  Gao, Spencer Williams, Sunny Gandhi, Tom Zur, Varun Iyer, and Zehua Li. 2023.
\newblock Legalbench: A collaboratively built benchmark for measuring legal
  reasoning in large language models.
\newblock In \emph{Proceedings of the 36th Conference on Neural Information
  Processing Systems (NeurIPS)}.

\bibitem[{Hendrycks et~al.(2021)Hendrycks, Burns, Chen, and
  Ball}]{hendrycks2021cuad}
Dan Hendrycks, Collin Burns, Anya Chen, and Spencer Ball. 2021.
\newblock Cuad: An expert-annotated nlp dataset for legal contract review.
\newblock \emph{NeurIPS}.

\bibitem[{Holzenberger et~al.(2020)Holzenberger, Blair-Stanek, and
  Durme}]{Holzenberger2020ADF}
Nils Holzenberger, Andrew Blair-Stanek, and Benjamin~Van Durme. 2020.
\newblock \href {https://api.semanticscholar.org/CorpusID:218581117} {A dataset
  for statutory reasoning in tax law entailment and question answering}.
\newblock In \emph{NLLP@KDD}.

\bibitem[{Hu et~al.(2018)Hu, Li, Tu, Liu, and Sun}]{hu-etal-2018-shot}
Zikun Hu, Xiang Li, Cunchao Tu, Zhiyuan Liu, and Maosong Sun. 2018.
\newblock \href {https://aclanthology.org/C18-1041} {Few-shot charge prediction
  with discriminative legal attributes}.
\newblock In \emph{Proceedings of the 27th International Conference on
  Computational Linguistics}, pages 487--498, Santa Fe, New Mexico, USA.
  Association for Computational Linguistics.

\bibitem[{Hwang et~al.(2024)Hwang, Lee, Cho, Lee, and
  Seo}]{10.5555/3600270.3602628}
Wonseok Hwang, Dongjun Lee, Kyoungyeon Cho, Hanuhl Lee, and Minjoon Seo. 2024.
\newblock A multi-task benchmark for korean legal language understanding and
  judgement prediction.
\newblock In \emph{Proceedings of the 36th International Conference on Neural
  Information Processing Systems}, NIPS '22, Red Hook, NY, USA. Curran
  Associates Inc.

\bibitem[{II et~al.(2021)II, Katz, and
  Detterman}]{Chapter11LexNLPNaturallanguageprocessingandinformationextractionforlegalandregulatorytexts}
Michael J.~Bommarito II, Daniel~Martin Katz, and Eric~M. Detterman. 2021.
\newblock \href {https://doi.org/10.4337/9781788972826.00017} {\emph{Chapter
  11: LexNLP: Natural language processing and information extraction for legal
  and regulatory texts}}.
\newblock Edward Elgar Publishing, Cheltenham, UK.

\bibitem[{Jin et~al.(2020)Jin, Guo, Qiu, and Zhang}]{jin-etal-2020-genwiki}
Zhijing Jin, Qipeng Guo, Xipeng Qiu, and Zheng Zhang. 2020.
\newblock \href {https://doi.org/10.18653/v1/2020.coling-main.217}
  {{G}en{W}iki: A dataset of 1.3 million content-sharing text and graphs for
  unsupervised graph-to-text generation}.
\newblock In \emph{Proceedings of the 28th International Conference on
  Computational Linguistics}, pages 2398--2409, Barcelona, Spain (Online).
  International Committee on Computational Linguistics.

\bibitem[{Katz et~al.(2017)Katz, Bommarito, and
  Blackman}]{10.1371/journal.pone.0174698}
Daniel~Martin Katz, Michael~J. Bommarito, II, and Josh Blackman. 2017.
\newblock \href {https://doi.org/10.1371/journal.pone.0174698} {A general
  approach for predicting the behavior of the supreme court of the united
  states}.
\newblock \emph{PLOS ONE}, 12(4):1--18.

\bibitem[{Katz et~al.(2023)Katz, Hartung, Gerlach, Jana, and
  Bommarito}]{Katz2023NaturalLP}
Daniel~Martin Katz, Dirk Hartung, Lauritz Gerlach, Abhik Jana, and
  Michael~James Bommarito. 2023.
\newblock \href {https://api.semanticscholar.org/CorpusID:256440319} {Natural
  language processing in the legal domain}.
\newblock \emph{ArXiv}, abs/2302.12039.

\bibitem[{Kaur and Bozic(2019)}]{Kaur2019ConvolutionalNN}
Arshdeep Kaur and Bojan Bozic. 2019.
\newblock \href {https://api.semanticscholar.org/CorpusID:207824257}
  {Convolutional neural network-based automatic prediction of judgments of the
  european court of human rights}.
\newblock In \emph{Irish Conference on Artificial Intelligence and Cognitive
  Science}.

\bibitem[{Koncel-Kedziorski et~al.(2019)Koncel-Kedziorski, Bekal, Luan, Lapata,
  and Hajishirzi}]{koncel-kedziorski-etal-2019-text}
Rik Koncel-Kedziorski, Dhanush Bekal, Yi~Luan, Mirella Lapata, and Hannaneh
  Hajishirzi. 2019.
\newblock \href {https://doi.org/10.18653/v1/N19-1238} {{T}ext {G}eneration
  from {K}nowledge {G}raphs with {G}raph {T}ransformers}.
\newblock In \emph{Proceedings of the 2019 Conference of the North {A}merican
  Chapter of the Association for Computational Linguistics: Human Language
  Technologies, Volume 1 (Long and Short Papers)}, pages 2284--2293,
  Minneapolis, Minnesota. Association for Computational Linguistics.

\bibitem[{Kung et~al.(2023)Kung, Cheatham, Medenilla, Sillos, De~Leon,
  Elepa{\~n}o, Madriaga, Aggabao, Diaz-Candido, Maningo
  et~al.}]{kung2023performance}
Tiffany~H Kung, Morgan Cheatham, Arielle Medenilla, Czarina Sillos, Lorie
  De~Leon, Camille Elepa{\~n}o, Maria Madriaga, Rimel Aggabao, Giezel
  Diaz-Candido, James Maningo, et~al. 2023.
\newblock Performance of chatgpt on usmle: potential for ai-assisted medical
  education using large language models.
\newblock \emph{PLoS digital health}, 2(2):e0000198.

\bibitem[{Lu et~al.(2022)Lu, Mishra, Xia, Qiu, Chang, Zhu, Tafjord, Clark, and
  Kalyan}]{lu2022learn}
Pan Lu, Swaroop Mishra, Tony Xia, Liang Qiu, Kai-Wei Chang, Song-Chun Zhu,
  Oyvind Tafjord, Peter Clark, and Ashwin Kalyan. 2022.
\newblock Learn to explain: Multimodal reasoning via thought chains for science
  question answering.
\newblock In \emph{The 36th Conference on Neural Information Processing Systems
  (NeurIPS)}.

\bibitem[{Luz~de Araujo et~al.(2018)Luz~de Araujo, de~Campos, de~Oliveira,
  Stauffer, Couto, and Bermejo}]{10.1007/978-3-319-99722-3_32}
Pedro~Henrique Luz~de Araujo, Te{\'o}filo~E. de~Campos, Renato R.~R.
  de~Oliveira, Matheus Stauffer, Samuel Couto, and Paulo Bermejo. 2018.
\newblock Lener-br: A dataset for named entity recognition in brazilian legal
  text.
\newblock In \emph{Computational Processing of the Portuguese Language}, pages
  313--323, Cham. Springer International Publishing.

\bibitem[{Medvedeva et~al.(2020)Medvedeva, Vols, and
  Wieling}]{Medvedeva2020-MEDUML}
Masha Medvedeva, Michel Vols, and Martijn Wieling. 2020.
\newblock \href {https://doi.org/10.1007/s10506-019-09255-y} {Using machine
  learning to predict decisions of the european court of human rights}.
\newblock \emph{Artificial Intelligence and Law}, 28(2):237--266.

\bibitem[{Niklaus et~al.(2021)Niklaus, Chalkidis, and
  St{\"u}rmer}]{niklaus-etal-2021-swiss}
Joel Niklaus, Ilias Chalkidis, and Matthias St{\"u}rmer. 2021.
\newblock \href {https://doi.org/10.18653/v1/2021.nllp-1.3}
  {{S}wiss-judgment-prediction: A multilingual legal judgment prediction
  benchmark}.
\newblock In \emph{Proceedings of the Natural Legal Language Processing
  Workshop 2021}, pages 19--35, Punta Cana, Dominican Republic. Association for
  Computational Linguistics.

\bibitem[{Niklaus et~al.(2023)Niklaus, Matoshi, Rani, Galassi, St{\"u}rmer, and
  Chalkidis}]{niklaus-etal-2023-lextreme}
Joel Niklaus, Veton Matoshi, Pooja Rani, Andrea Galassi, Matthias St{\"u}rmer,
  and Ilias Chalkidis. 2023.
\newblock \href {https://doi.org/10.18653/v1/2023.findings-emnlp.200}
  {{LEXTREME}: A multi-lingual and multi-task benchmark for the legal domain}.
\newblock In \emph{Findings of the Association for Computational Linguistics:
  EMNLP 2023}, pages 3016--3054, Singapore. Association for Computational
  Linguistics.

\bibitem[{OpenAI(2023)}]{OpenAI2023GPT4TR}
OpenAI. 2023.
\newblock {GPT}-4 technical report.
\newblock Technical report.

\bibitem[{Pais et~al.(2021)Pais, Mitrofan, Gasan, Coneschi, and
  Ianov}]{pais-etal-2021-named}
Vasile Pais, Maria Mitrofan, Carol~Luca Gasan, Vlad Coneschi, and Alexandru
  Ianov. 2021.
\newblock \href {https://doi.org/10.18653/v1/2021.nllp-1.2} {Named entity
  recognition in the {R}omanian legal domain}.
\newblock In \emph{Proceedings of the Natural Legal Language Processing
  Workshop 2021}, pages 9--18, Punta Cana, Dominican Republic. Association for
  Computational Linguistics.

\bibitem[{Riviere et~al.(2024)Riviere, Pathak, Sessa, Hardin, Bhupatiraju,
  Hussenot, Mesnard, Shahriari, Ram'e, Ferret, Liu, Tafti, Friesen, Casbon,
  Ramos, Kumar, Lan, Jerome, Tsitsulin, Vieillard, Stańczyk, Girgin, Momchev,
  Hoffman, Thakoor, Grill, Neyshabur, Walton, Severyn, Parrish, Ahmad,
  Hutchison, Abdagic, Carl, Shen, Brock, Coenen, Laforge, Paterson, Bastian,
  Piot, Wu, Royal, Chen, Kumar, Perry, Welty, Choquette-Choo, Sinopalnikov,
  Weinberger, Vijaykumar, Rogozi'nska, Herbison, Bandy, Wang, Noland, Moreira,
  Senter, Eltyshev, Visin, Rasskin, Wei, Cameron, Martins, Hashemi,
  Klimczak-Pluci'nska, Batra, Dhand, Nardini, Mein, Zhou, Svensson, Stanway,
  Chan, Zhou, Carrasqueira, Iljazi, Becker, Fernandez, van Amersfoort, Gordon,
  Lipschultz, Newlan, Ji, Mohamed, Badola, Black, Millican, McDonell, Nguyen,
  Sodhia, Greene, Sjoesund, Usui, Sifre, Heuermann, Lago, McNealus, Soares,
  Kilpatrick, Dixon, Martins, Reid, Singh, Iverson, Gorner, Velloso, Wirth,
  Davidow, Miller, Rahtz, Watson, Risdal, Kazemi, Moynihan, Zhang, Kahng, Park,
  Rahman, Khatwani, Dao, Bardoliwalla, Devanathan, Dumai, Chauhan, Wahltinez,
  Botarda, Barnes, Barham, Michel, Jin, Georgiev, Culliton, Kuppala, Comanescu,
  Merhej, Jana, Rokni, Agarwal, Mullins, Saadat, Carthy, Perrin, Arnold,
  Krause, Dai, Garg, Sheth, Ronstrom, Chan, Jordan, Yu, Eccles, Hennigan,
  Kocisk{\'y}, Doshi, Jain, Yadav, Meshram, Dharmadhikari, Barkley, Wei, Ye,
  Han, Kwon, Xu, Shen, Gong, Wei, Cotruta, Kirk, Rao, Giang, Peran, Warkentin,
  Collins, Barral, Ghahramani, Hadsell, Sculley, Banks, Dragan, Petrov,
  Vinyals, Dean, Hassabis, Kavukcuoglu, Farabet, Buchatskaya, Borgeaud, Fiedel,
  Joulin, Kenealy, Dadashi, and Andreev}]{Riviere2024Gemma2I}
Gemma Team~Morgane Riviere, Shreya Pathak, Pier~Giuseppe Sessa, Cassidy Hardin,
  Surya Bhupatiraju, L'eonard Hussenot, Thomas Mesnard, Bobak Shahriari,
  Alexandre Ram'e, Johan Ferret, Peter Liu, Pouya~Dehghani Tafti, Abe Friesen,
  Michelle Casbon, Sabela Ramos, Ravin Kumar, Charline~Le Lan, Sammy Jerome,
  Anton Tsitsulin, Nino Vieillard, Piotr Stańczyk, Sertan Girgin, Nikola
  Momchev, Matt Hoffman, Shantanu Thakoor, Jean-Bastien Grill, Behnam
  Neyshabur, Alanna Walton, Aliaksei Severyn, Alicia Parrish, Aliya Ahmad,
  Allen Hutchison, Alvin Abdagic, Amanda Carl, Amy Shen, Andy Brock, Andy
  Coenen, Anthony Laforge, Antonia Paterson, Ben Bastian, Bilal Piot, Boxi Wu,
  Brandon Royal, Charlie Chen, Chintu Kumar, Chris Perry, Christoper~A. Welty,
  Christopher~A. Choquette-Choo, Danila Sinopalnikov, David Weinberger, Dimple
  Vijaykumar, Dominika Rogozi'nska, D.~Herbison, Elisa Bandy, Emma Wang, Eric
  Noland, Erica Moreira, Evan Senter, Evgenii Eltyshev, Francesco Visin,
  Gabriel Rasskin, Gary Wei, Glenn Cameron, Gus Martins, Hadi Hashemi, Hanna
  Klimczak-Pluci'nska, Harleen Batra, Harsh Dhand, Ivan Nardini, Jacinda Mein,
  Jack Zhou, James Svensson, Jeff Stanway, Jetha Chan, Jin Zhou, Joana
  Carrasqueira, Joana Iljazi, Jocelyn Becker, Joe Fernandez, Joost~R. van
  Amersfoort, Josh Gordon, Josh Lipschultz, Joshua Newlan, Junsong Ji, Kareem
  Mohamed, Kartikeya Badola, Kat Black, Katie Millican, Keelin McDonell, Kelvin
  Nguyen, Kiranbir Sodhia, Kish Greene, Lars~Lowe Sjoesund, Lauren Usui,
  L.~Sifre, L.~Heuermann, Leticia Lago, Lilly McNealus, Livio~Baldini Soares,
  Logan Kilpatrick, Lucas Dixon, Luciano Martins, Machel Reid, Manvinder Singh,
  Mark Iverson, Martin Gorner, Mat Velloso, Mateo Wirth, Matt Davidow, Matt
  Miller, Matthew Rahtz, Matthew Watson, Meg Risdal, Mehran Kazemi, Michael
  Moynihan, Ming Zhang, Minsuk Kahng, Minwoo Park, Mofi Rahman, Mohit Khatwani,
  Natalie Dao, Nenshad Bardoliwalla, Nesh Devanathan, Neta Dumai, Nilay
  Chauhan, Oscar Wahltinez, Pankil Botarda, Parker Barnes, Paul Barham, Paul
  Michel, Pengchong Jin, Petko Georgiev, Phil Culliton, Pradeep Kuppala, Ramona
  Comanescu, Ramona Merhej, Reena Jana, Reza Rokni, Rishabh Agarwal, Ryan
  Mullins, Samaneh Saadat, S.~Mc Carthy, Sarah Perrin, S'ebastien Arnold,
  Sebastian Krause, Shengyang Dai, Shruti Garg, Shruti Sheth, Sue Ronstrom,
  Susan Chan, Timothy Jordan, Ting Yu, Tom Eccles, Tom Hennigan, Tom{\'a}s
  Kocisk{\'y}, Tulsee Doshi, Vihan Jain, Vikas Yadav, Vilobh Meshram, Vishal
  Dharmadhikari, Warren Barkley, Wei Wei, Wenming Ye, Woohyun Han, Woosuk Kwon,
  Xiang Xu, Zhe Shen, Zhitao Gong, Zichuan Wei, Victor Cotruta, Phoebe Kirk,
  Anand Rao, Minh Giang, Ludovic Peran, Tris~Brian Warkentin, Eli Collins,
  Joelle Barral, Zoubin Ghahramani, Raia Hadsell, D.~Sculley, Jeanine Banks,
  Anca Dragan, Slav Petrov, Oriol Vinyals, Jeffrey Dean, Demis Hassabis, Koray
  Kavukcuoglu, Cl'ement Farabet, Elena Buchatskaya, Sebastian Borgeaud, Noah
  Fiedel, Armand Joulin, Kathleen Kenealy, Robert Dadashi, and Alek Andreev.
  2024.
\newblock \href {https://api.semanticscholar.org/CorpusID:270843326} {Gemma 2:
  Improving open language models at a practical size}.
\newblock \emph{ArXiv}, abs/2408.00118.

\bibitem[{Rozi{\`e}re et~al.(2023)Rozi{\`e}re, Gehring, Gloeckle, Sootla, Gat,
  Tan, Adi, Liu, Remez, Rapin, Kozhevnikov, Evtimov, Bitton, Bhatt, Ferrer,
  Grattafiori, Xiong, D'efossez, Copet, Azhar, Touvron, Martin, Usunier,
  Scialom, and Synnaeve}]{Rozire2023CodeLO}
Baptiste Rozi{\`e}re, Jonas Gehring, Fabian Gloeckle, Sten Sootla, Itai Gat,
  Xiaoqing Tan, Yossi Adi, Jingyu Liu, Tal Remez, J{\'e}r{\'e}my Rapin, Artyom
  Kozhevnikov, I.~Evtimov, Joanna Bitton, Manish~P Bhatt, Cristian~Cant{\'o}n
  Ferrer, Aaron Grattafiori, Wenhan Xiong, Alexandre D'efossez, Jade Copet,
  Faisal Azhar, Hugo Touvron, Louis Martin, Nicolas Usunier, Thomas Scialom,
  and Gabriel Synnaeve. 2023.
\newblock \href {https://api.semanticscholar.org/CorpusID:261100919} {Code
  llama: Open foundation models for code}.
\newblock \emph{ArXiv}, abs/2308.12950.

\bibitem[{Shi et~al.(2022)Shi, Fried, Ghazvininejad, Zettlemoyer, and
  Wang}]{shi-etal-2022-natural}
Freda Shi, Daniel Fried, Marjan Ghazvininejad, Luke Zettlemoyer, and Sida~I.
  Wang. 2022.
\newblock \href {https://doi.org/10.18653/v1/2022.emnlp-main.231} {Natural
  language to code translation with execution}.
\newblock In \emph{Proceedings of the 2022 Conference on Empirical Methods in
  Natural Language Processing}, pages 3533--3546, Abu Dhabi, United Arab
  Emirates. Association for Computational Linguistics.

\bibitem[{Tuggener et~al.(2020)Tuggener, von D{\"a}niken, Peetz, and
  Cieliebak}]{tuggener-etal-2020-ledgar}
Don Tuggener, Pius von D{\"a}niken, Thomas Peetz, and Mark Cieliebak. 2020.
\newblock \href {https://aclanthology.org/2020.lrec-1.155} {{LEDGAR}: A
  large-scale multi-label corpus for text classification of legal provisions in
  contracts}.
\newblock In \emph{Proceedings of the Twelfth Language Resources and Evaluation
  Conference}, pages 1235--1241, Marseille, France. European Language Resources
  Association.

\bibitem[{Urchs. et~al.(2021)Urchs., Mitrović., and Granitzer.}]{icaart21}
Stefanie Urchs., Jelena Mitrović., and Michael Granitzer. 2021.
\newblock \href {https://doi.org/10.5220/0010187305150521} {Design and
  implementation of german legal decision corpora}.
\newblock In \emph{Proceedings of the 13th International Conference on Agents
  and Artificial Intelligence - Volume 2: ICAART}, pages 515--521. INSTICC,
  SciTePress.

\bibitem[{Xiao et~al.(2018)Xiao, Zhong, Guo, Tu, Liu, Sun, Feng, Han, Hu, Wang,
  and Xu}]{Xiao2018CAIL2018AL}
Chaojun Xiao, Haoxiang Zhong, Zhipeng Guo, Cunchao Tu, Zhiyuan Liu, Maosong
  Sun, Yansong Feng, Xianpei Han, Zhen Hu, Heng Wang, and Jianfeng Xu. 2018.
\newblock \href {https://api.semanticscholar.org/CorpusID:49652844} {Cail2018:
  A large-scale legal dataset for judgment prediction}.
\newblock \emph{ArXiv}, abs/1807.02478.

\bibitem[{Ye et~al.(2018)Ye, Jiang, Luo, and Chao}]{ye-etal-2018-interpretable}
Hai Ye, Xin Jiang, Zhunchen Luo, and Wenhan Chao. 2018.
\newblock \href {https://doi.org/10.18653/v1/N18-1168} {Interpretable charge
  predictions for criminal cases: Learning to generate court views from fact
  descriptions}.
\newblock In \emph{Proceedings of the 2018 Conference of the North {A}merican
  Chapter of the Association for Computational Linguistics: Human Language
  Technologies, Volume 1 (Long Papers)}, pages 1854--1864, New Orleans,
  Louisiana. Association for Computational Linguistics.

\bibitem[{Zala et~al.(2023)Zala, Lin, Cho, and Bansal}]{Zala2023DiagrammerGPT}
Abhay Zala, Han Lin, Jaemin Cho, and Mohit Bansal. 2023.
\newblock Diagrammergpt: Generating open-domain, open-platform diagrams via llm
  planning.

\bibitem[{Zhang et~al.(2020)Zhang, Kishore, Wu, Weinberger, and
  Artzi}]{bertscore}
Tianyi Zhang, Varsha Kishore, Felix Wu, Kilian~Q. Weinberger, and Yoav Artzi.
  2020.
\newblock \href {https://openreview.net/forum?id=SkeHuCVFDr} {Bertscore:
  Evaluating text generation with {BERT}}.
\newblock In \emph{8th International Conference on Learning Representations,
  {ICLR} 2020, Addis Ababa, Ethiopia, April 26-30, 2020}. OpenReview.net.

\bibitem[{Zheng et~al.(2021)Zheng, Guha, Anderson, Henderson, and
  Ho}]{10.1145/3462757.3466088}
Lucia Zheng, Neel Guha, Brandon~R. Anderson, Peter Henderson, and Daniel~E. Ho.
  2021.
\newblock \href {https://doi.org/10.1145/3462757.3466088} {When does
  pretraining help? assessing self-supervised learning for law and the casehold
  dataset of 53,000+ legal holdings}.
\newblock In \emph{Proceedings of the Eighteenth International Conference on
  Artificial Intelligence and Law}, ICAIL '21, page 159–168, New York, NY,
  USA. Association for Computing Machinery.

\end{thebibliography}
